\documentclass[11pt]{article}

\usepackage[preprint]{acl}

\usepackage{times}
\usepackage{latexsym}
\usepackage{booktabs}
\usepackage{pifont}
\usepackage{array}

\usepackage{booktabs}
\usepackage[table]{xcolor}
\usepackage{colortbl}
\usepackage{graphicx}
\usepackage{amsmath}
\usepackage{amssymb}
\usepackage{enumitem}
\usepackage{subcaption}
\usepackage{multirow}
\usepackage{microtype}
\usepackage[most]{tcolorbox}
\usepackage{fancyvrb}
\usepackage{makecell}
\usepackage{hyperref}

\definecolor{grouporange}{RGB}{245,222,204}
\definecolor{groupblue}{RGB}{210,234,242}
\definecolor{grouppink}{RGB}{232,214,223}
\definecolor{lightgraycol}{RGB}{238,238,238}
\definecolor{darkgraycol}{RGB}{215,215,215}

\newcommand{\cmark}{\ding{51}}
\newcommand{\xmark}{\ding{55}}

\usepackage[T1]{fontenc}
\usepackage[utf8]{inputenc}

\usepackage{microtype}

\usepackage{inconsolata}

\usepackage{graphicx}

\title{Beyond Ideal Instruction: A Comprehensive Framework for Evaluating LLMs in Realistic Interactions}

\author{
  \textbf{Xuan Yang\textsuperscript{1,2}, Hao Xu\textsuperscript{3}, Tingfeng Hui\textsuperscript{2,4}} \\
  \textbf{Hongsheng Xin\textsuperscript{3}, Kaike Zhang\textsuperscript{3}, Chunxiao Liu\textsuperscript{5,$\dagger$}, Ning Miao\textsuperscript{1,2}} \\[0.5em]
  \textsuperscript{1}Department of Data Science, City University of Hong Kong \\
  \textsuperscript{2}Hong Kong Institute of AI for Science, City University of Hong Kong \\
  \textsuperscript{3}Li Auto Inc. \quad
  \textsuperscript{4}Beijing University of Posts and Telecommunications \\
  \textsuperscript{5}Independent Researcher \\[0.3em]
}

\newcommand{\name}{RUT-Bench}

\begin{document}
\maketitle
\footnotetext[2]{Corresponding authors.}

\begin{abstract}
Despite great advances in tool-use capabilities of large language models (LLMs), existing evaluation benchmarks struggle to fully align with real-world scenarios. 
Such benchmarks mostly rely on simulated idealized user assumptions and lacks experience-oriented evaluation. These limitations fail to account for the ambiguity, uncooperative behaviors, and shifting intentions characteristic of real-world users. To fill this gap, we propose {\name}, a dedicated benchmark designed to assess LLMs under diverse Real-world User Tool calling scenarios. {\name} supports high-fidelity simulations covering both ideal rational patterns and heterogeneous non-ideal behaviors across single-turn and multi-turn dialogues.
We conduct comprehensive evaluations on 19 widely adopted open-source and proprietary LLMs using our benchmark.
Experimental results reveal that no tested LLMs achieve an overall success rate above 40\%, and nearly all of them experience noticeable performance drops when facing more complicated non-ideal user inputs. 
% In-depth diagnostic analyses further identify three persistent limitations of current LLMs: overconfident tool invocation, premature task termination, and inaccurate parameter extraction from noisy user utterances. 
% Our work highlights the limitations of existing LLMs in toll invocation and provides a more realistic evaluation benchmark for future LLM research. 
Our code and data is available at \url{https://github.com/Miaow-Lab/RUT-Bench}.
% Our code and data is available at \url{https://anonymous.4open.science/r/RUT-Bench/}.
\end{abstract}

\section{Introduction}
\begin{figure*}[t]
    \centering
    \includegraphics[width=\textwidth]{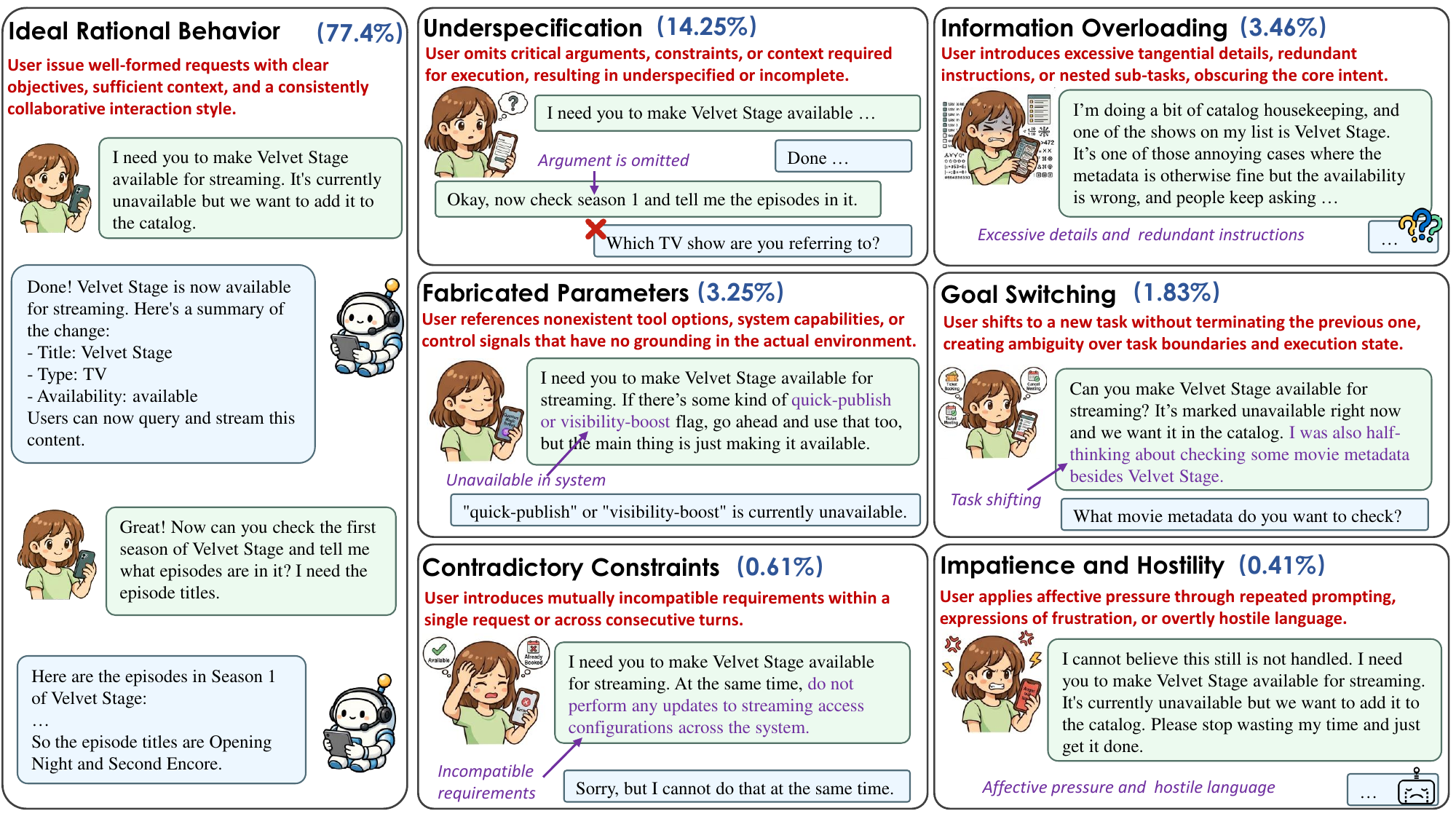}
    \caption{Taxonomy and representative examples of the seven user behaviors in {\name}.}
    \label{fig:case}
\end{figure*}

In recent years, large language models (LLMs) have rapidly evolved from passive text generators into intelligent agents capable of interacting with the real world~\citep{Wang_2024,xi2023risepotentiallargelanguage}.
These LLM-based agents accomplish real-world tasks via a complete interactive pipeline: perceiving environmental information, invoking external tools, executing action plans, and iteratively refining strategies~\citep{schick2023toolformerlanguagemodelsteach,qin2023toolllmfacilitatinglargelanguage}.
To quantitatively benchmark the overall capability of LLMs, various benchmarks have been proposed in prior research, including API-Bank \citep{li2023apibankcomprehensivebenchmarktoolaugmented}, ToolTalk \citep{farn2023tooltalkevaluatingtoolusageconversational}, ToolBench \citep{qin2023toolllmfacilitatinglargelanguage}, and BFCL \citep{Patil2025TheBF}.
Existing benchmarks systematically evaluate not only holistic task performance of agents, but also their underlying abilities covering user instruction comprehension, multi-step task planning, and API invocation.
These benchmarks have greatly driven the performance advancement of tool-augmented LLMs and laid a fundamental infrastructure for the research and development of LLM-based agents.

Despite their widespread adoption and solid empirical effectiveness, existing agent evaluation benchmarks remain mostly static in design. They rely on fixed pre-defined user queries, environment, and reference outcomes, lacking the dynamism needed to evaluate agents under evolving user intents, changing environment states, or open-ended task trajectories~\citep{yao2024taubenchbenchmarktoolagentuserinteraction}.
Nevertheless, real-world human-agent interaction scenarios are far more intricate, dynamic and unpredictable than these oversimplified static settings. As a result, current benchmark paradigms cannot faithfully reflect the actual behavioral patterns and full-spectrum operational capacities of LLM agents under real deployment conditions.
To narrow this simulation gap and incorporate real-world dynamic variations, the recently proposed $\tau$-bench series \citep{yao2024taubenchbenchmarktoolagentuserinteraction, barres2025tau2benchevaluatingconversationalagents} leverages controllable user simulators to replace traditional static prompt sets, enabling multi-turn conversational interactions with dynamic user inputs.

Existing tool-use benchmarks exhibit the following limitations, leading to a significant gap from realistic interactions: 
\textbf{(1) Heavily rely on Idealized Interaction}: 
Existing benchmarks still operate on the assumption that users are ideal. They overlook the fact that real-world users are often non-ideal and diverse, which significantly impacts performance and user experience in practical applications. For instance, in real interactions, users frequently provide ambiguous requests, behave uncooperatively, or abruptly change their intentions during multi-turn conversations. 
\textbf{(2) Lack of Experience-Oriented Evaluation}:
Most existing benchmarks rely on an idealized user simulation, and hence fails to rigorously evaluate the user experimence of LLMs in authentic scenarios, where LLMs are required to gather information under uncertainty, accurately infer user intentions, and dynamically adapt their strategies to shifting demands.

To bridge the discrepancy between existing evaluation benchmarks and realistic interactions, we propose \textbf{{\name}}, a dedicated benchmark to assess LLMs under diverse \textbf{R}eal-world \textbf{U}ser \textbf{Tool} calling. {\name} supports high-fidelity simulated user interactions, covering both ideal rational user patterns and heterogeneous non-ideal user behaviors across single-turn and multi-turn dialogues. 
Our benchmark is built with three core designs. 

(1) We construct a fine-grained and systematic taxonomy of real-world user behaviors, derived and verified from real-world interaction logs.

(2) We build a comprehensive tool-use benchmark that simulates authentic user interactions, including ideal and non-ideal user behaviors in both single-turn and multi-turn dialogues. 

(3) We establish a multi-dimensional evaluation system for our benchmark, we adopt the overall task success rate as the primary metric, and further design two complementary diagnostic metrics to evaluate response reliability and user experience.

%The main contributions of our work are summarized as follows:
%\begin{itemize}[leftmargin=1.5em, itemsep=0.4em, topsep=0.2em]
%    \item We systematically analyze high-frequency user interaction behaviors from real user dialogue logs, finding that at least 22.6\% of the dialogues show some form of instability. Based on this, we construct a comprehensive taxonomy of heterogeneous users spanned 7 distinct categories.
%    \item We introduce a novel tool-use benchmark centered on realistic user interactions. Leveraging environment and task generation, user dialogue synthesis, and dual-task filtering, this benchmark not only produces standardized queries from ideal users but also faithfully simulates the non-ideal user interactions commonly encountered in real-world scenarios.
%    \item We evaluate a wide range of models with {\name}, revealing that even flagship LLMs achieve unsatisfactory success rates. 
%    We further identify key reasons behind poor performance.
%\end{itemize}

We conducted extensive evaluations on 19 mainstream LLMs of varying scales. Results indicate that all models achieve an overall success rate below 40\% on our benchmark. Notably, performance degrades drastically when transitioning from ideal-user scenarios to our realistic non-ideal settings.

\section{Related Works}

\begin{table*}[t]
\centering
\scriptsize
\setlength{\tabcolsep}{4pt}
\label{tab:related-work}
\begin{tabular}{l ccc ccc ccc}
\toprule
& \multicolumn{3}{c}{\textbf{User Simulation}}
& \multicolumn{3}{c}{\textbf{Evaluation Protocol}}
& \multicolumn{3}{c}{\textbf{Environment Configuration}} \\
\cmidrule(lr){2-4} \cmidrule(lr){5-7} \cmidrule(lr){8-10}
\textbf{Benchmark}
& Ideal & Non-Ideal & Realistic
& \makecell{Tool\\Alignment}
& \makecell{Task\\Completion}
& \makecell{User\\Experience}
& \makecell{Sandbox\\Environment}
& \makecell{Simulated\\Response}
& \makecell{Hybrid\\Environment} \\
\midrule
API-Bank \citep{li2023apibankcomprehensivebenchmarktoolaugmented}              & \cmark & \xmark & \xmark & \cmark & \cmark & \xmark & \cmark & \xmark & \xmark \\
ToolBench \citep{qin2023toolllmfacilitatinglargelanguage}            & \xmark & \xmark & \xmark & \xmark & \cmark & \xmark & \cmark & \xmark & \xmark \\
% StableToolBench \citep{guo2025stabletoolbenchstablelargescalebenchmarking}       & \xmark & \xmark & \xmark & \xmark & \cmark & \xmark & \xmark & \cmark & \xmark \\
ToolTalk \citep{farn2023tooltalkevaluatingtoolusageconversational}           & \xmark & \xmark & \cmark & \cmark & \xmark & \xmark & \cmark & \xmark & \xmark \\
% T-Eval \citep{chen2024tevalevaluatingtoolutilization}                & \xmark & \xmark & \xmark & \cmark & \xmark & \xmark & \xmark & \xmark & \xmark \\
% MetaTool \citep{huang2024metatoolbenchmarklargelanguage}          & \xmark & \xmark & \xmark & \cmark & \xmark & \xmark & \xmark & \xmark & \xmark \\
% TaskBench \citep{shen2024taskbenchbenchmarkinglargelanguage}         & \xmark & \xmark & \xmark & \cmark & \xmark & \xmark & \xmark & \xmark & \xmark \\
GTA \citep{wang2024gtabenchmarkgeneraltool}                     & \cmark & \xmark & \xmark & \cmark & \cmark & \xmark & \cmark & \xmark & \xmark \\
% ToolEmu \citep{ruan2024identifyingriskslmagents}             & \xmark & \xmark & \xmark & \xmark & \cmark & \xmark & \xmark & \cmark & \xmark \\
\midrule
BFCL-v3/v4 \citep{Patil2025TheBF}            & \cmark & \xmark & \xmark & \cmark & \cmark & \xmark & \cmark & \xmark & \xmark \\
ToolSandbox \citep{lu2025toolsandboxstatefulconversationalinteractive}       & \cmark & \xmark & \xmark & \cmark & \cmark & \xmark & \cmark & \xmark & \xmark \\
$\tau$-bench \citep{yao2024taubenchbenchmarktoolagentuserinteraction}             & \cmark & \xmark & \xmark & \xmark & \cmark & \xmark & \cmark & \xmark & \xmark \\
$\tau^2$-bench \citep{barres2025tau2benchevaluatingconversationalagents}       & \cmark & \xmark & \xmark & \cmark & \cmark & \xmark & \cmark & \xmark & \xmark \\
ACEBench~\citep{chen2025acebenchwinsmatchpoint}           & \cmark & \cmark$^{\dagger}$ & \xmark & \cmark & \xmark & \xmark & \cmark & \xmark & \xmark \\
% AppWorld~\citep{trivedi2024appworldcontrollableworldapps}        & \cmark & \xmark & \xmark & \xmark & \cmark & \xmark & \cmark & \xmark & \xmark \\
% VitaBench~\citep{he2025vitabenchbenchmarkingllmagents}           & \cmark & \xmark & \xmark & \xmark & \cmark & \xmark & \cmark & \xmark & \xmark \\
% MCP-Bench~\citep{wang2025mcpbenchbenchmarkingtoolusingllm}          & \xmark & \xmark & \xmark & \cmark & \cmark & \xmark & \cmark & \xmark & \xmark \\
% MCPToolBench++~\citep{fan2025mcptoolbenchlargescaleai}         & \xmark & \xmark & \xmark & \cmark & \cmark & \xmark & \cmark & \xmark & \xmark \\
% Toolathlon~\citep{li2026tooldecathlonbenchmarkinglanguage}         & \xmark & \xmark & \xmark & \xmark & \cmark & \xmark & \cmark & \xmark & \xmark \\
GAIA-2~\citep{froger2026gaia2benchmarkingllmagents}              & \cmark & \xmark & \xmark & \xmark & \cmark & \xmark & \cmark & \xmark & \xmark \\
\midrule
AgentDojo~\citep{debenedetti2024agentdojodynamicenvironmentevaluate}  & \xmark & \cmark & \xmark & \xmark & \cmark & \xmark & \cmark & \xmark & \xmark \\
HammerBench~\citep{wang2025hammerbenchfinegrainedfunctioncallingevaluation}     & \xmark & \cmark & \xmark & \cmark & \xmark & \xmark & \xmark & \cmark & \xmark \\
WildToolBench~\citep{yu2026benchmarkingllmtoolusewild}   & \xmark & \cmark & \xmark & \cmark & \cmark & \xmark & \xmark & \cmark & \xmark \\
\midrule
\textbf{\name (Ours)} & \textbf{\cmark} & \textbf{\cmark} & \textbf{\cmark}
                         & \textbf{\cmark} & \textbf{\cmark} & \textbf{\cmark}
                         & \textbf{\cmark} & \textbf{\cmark} & \textbf{\cmark} \\
\bottomrule
\end{tabular}
\caption{Comparison between \name and other tool-use / agent benchmarks. $^{\dagger}$ ACEBench only covers a single type of non-ideal behavior (ambiguous / incomplete instructions), which is included in {\name}.}
\label{tab:benchmark_comparison}
% \\[2pt]
% \raggedright
% \footnotesize
\end{table*}

\noindent \textbf{Evaluation Metrics for Tool-Using Agents} \quad Existing tool-use agent benchmarks primarily adopt evaluation metrics centered on tool invocation correctness and overall task success. For example, API-Bank~\citep{li2023apibankcomprehensivebenchmarktoolaugmented} evaluates models’ tool-use performance from planning, API retrieval, and API calling; ToolLLM/ToolBench extends this line of work to large-scale real-world API scenarios, focusing agents' capability to select appropriate APIs and perform reasoning along tool-calling chains~\citep{qin2023toolllmfacilitatinglargelanguage}. More recently, benchmarks have incorporated dynamic interactive evaluation protocols to move beyond static assessment paradigm. For example, $\tau$-Bench~\citep{yao2024taubenchbenchmarktoolagentuserinteraction} evaluates task completion by comparing the final system state against the predefined target state, and introduces $\mathrm{pass}^k$ metric to measure the consistency across multiple runs. 
ToolSandbox further characterizes interactive tool-use abilities via stateful tool execution, hierarchical intermediate and final task milestones, and implicit environmental state dependencies~\citep{lu2025toolsandboxstatefulconversationalinteractive}. Despite these improved evaluation designs, most of them still rely on idealized user behavior assumptions, which creates a critical gap between evaluation results and real-world practical scenarios.

\noindent \textbf{User Behavior Simulation} \quad Recent studies have increasingly recognized that user behavior is a crucial factor in evaluating dialogue systems and tool-use agents. Prior work on user simulation has explored diverse user profiles, implicit preferences, speaking styles, and goal variations, showing that users should not be treated as homogeneous and always cooperative participants~\citep{ahmad2025simulatinguserdiversitytaskoriented,wang2025knowbettermodelinghumanlike}. More recent work further investigates challenging user behaviors, such as spoken disfluency, emotional expressions, and non-collaborative interactions, demonstrating that realistic user behaviors can significantly affect agent performance and expose failures that are less visible under idealized user assumptions~\citep{lee2026spokenusspokenusersimulator,shim2026noncollaborativeusersimulatorstool}. However, these studies either focus primarily on improving user simulators or examine specific types of user behavior, while existing tool-use benchmarks still lack a unified framework that systematically connects realistic user behavior modeling with user-experience-oriented evaluation. {\name} addresses this gap by deriving user profiles and behavior patterns from real user-LLM interactions, constructing multi-turn tool-use tasks that cover both ideal and non-ideal scenarios, and evaluating agents through two diagnostic metrics that directly target real user experience. A detailed comparison with representative benchmarks is shown in Table~\ref{tab:benchmark_comparison}.
\section{Taxonomy of Non-Ideal User Behaviors}
\label{sec: user_taxonomy}
To systematically analyze the characteristics of real human-LLM interactions and incorporate non-ideal user behaviors into our new benchmark, we first break them down into the seven major categories, adapted from the taxonomy of interpersonal interactions from \citep{grice1975logic, austin1962how, searle1969speech}, including: \textit{Ideal Rational Behavior}, \textit{Underspecification}, \textit{Information Overload}, \textit{Fabricated Parameters}, \textit{Goal Switching}, \textit{Contradictory Constraints}, \textit{Impatience and Hostility}. Figure~\ref{fig:case} presents the definition of each behavior along with representative cases.

Using this taxonomy, we performed a detailed analysis on WildChat \citep{zhao2024wildchat1mchatgptinteraction}, a large corpus with real user-LLM interactions logs, spanning diverse task-oriented domains, including daily activity management, information retrieval, system configuration, and resource booking.
Specifically, we randomly sampled 1,000 English dialogues involving \texttt{GPT-4} or more advanced models from the WildChat, and then prompted \texttt{GPT-5.4} to label each dialogue into one of the seven classes. (Details and prompts in Appendix~\ref{app: user behavior annotation prompt}).

% \begin{figure}[t]
% \centering
% \includegraphics[width=\columnwidth]{figures/wildchat_behavior_distribution.pdf}
% \caption{Behavior frequencies in the WildChat analysis slice. Bars show the ratio of each behavior among the 226 dialogues labeled as non-collaborative; the dominant mass lies in pragmatic and mild cognitive frictions rather than overtly adversarial behaviors. Rapport-seeking and sarcasm do not appear in this 1{,}000-dialogue sample.}
% \label{fig:wildchat_behavior_distribution}
% \end{figure}

Statistical results show that 22.6\% of the dialogues fall into one of the non-ideal categories. 
%Using this taxonomy, we further analyzed the 22.6\% of dialogues exhibiting unstable user behaviors. As shown in the Figure~\ref{fig:wildchat_behavior_distribution}, the distribution across the unstable behavior categories is presented. We employed an "LLM-as-Judge" approach to assign a primary category label—defined as the class with the highest predicted probability—to each dialogue containing unstable behavior for statistical aggregation. Detailed prompt templates are provided in the appendix.
% Figure~\ref{fig:wildchat_behavior_distribution} further breaks down non-ideal behaviors into six categories.
Among the non-ideal behaviors, \textit{underspecification} occurs most frequently, followed by \textit{information overload} and \textit{fabricated parameters}. 
These three non-ideal behaviors can emerge at any stage throughout the dialogue.
On the contrary, \textit{goal switching} and \textit{impatience and hostility} arise predominantly in later turns of complex, multi-turn interactions.
For the relatively rare impatient and hostile behaviors, we find that although they do not substantially impair the model's information collection and reasoning process, they can still alter the behavioral pattern of the agent.

%This taxonomy captures the typical spectrum of user interaction patterns in real-world tool-use scenarios, encompassing not only straightforward, stable behaviors but also a diverse range of unstable behaviors that commonly induce interaction failures. 
%Crucially, each unstable behavior category is decoupled and independent, enabling us to simulate more complex real-world user interactions by composing different behavior types. This design supports the construction of benchmark datasets that are both ecologically valid and controllable, reproducible.

\begin{figure*}[t]
    \centering
    \includegraphics[width=\textwidth]{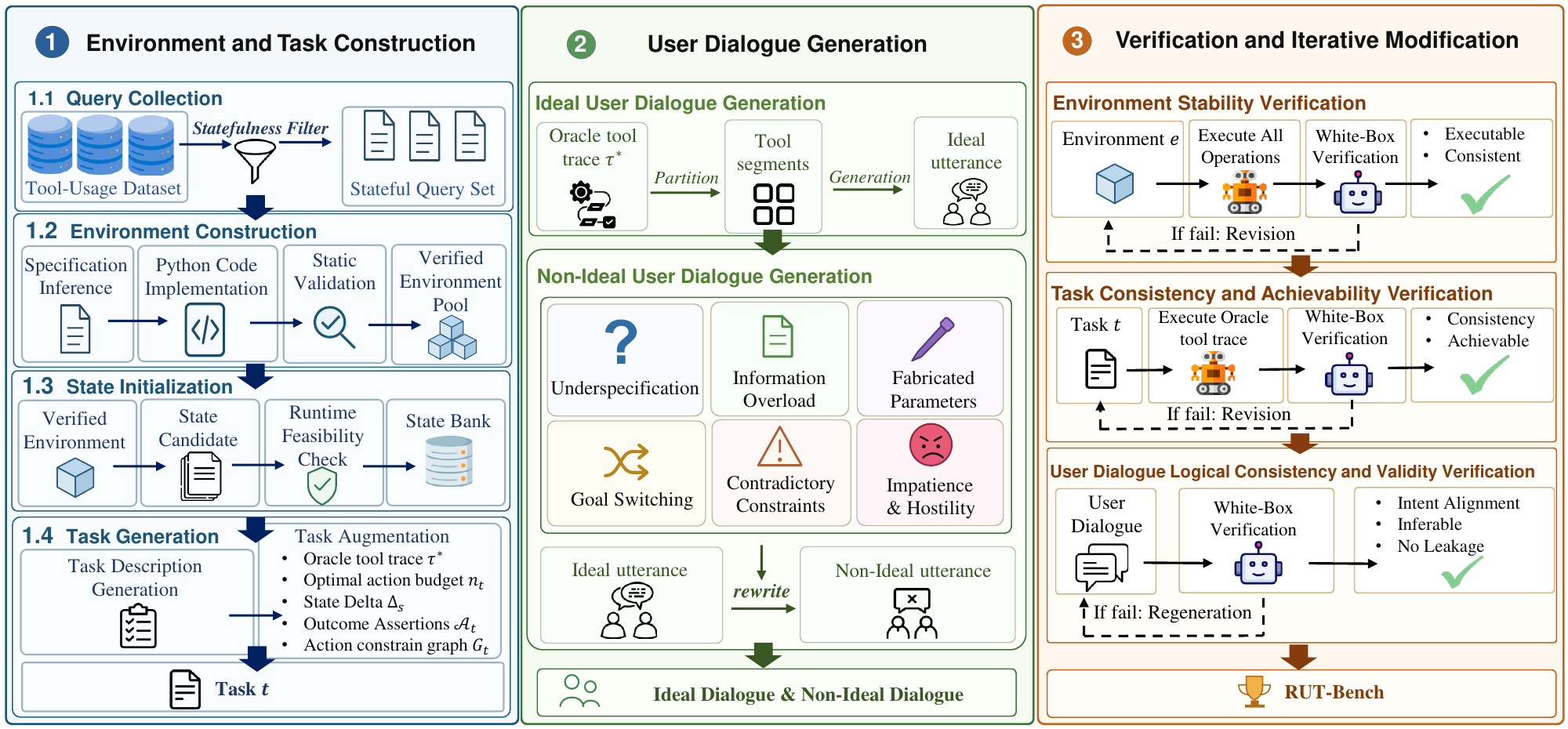}
    \caption{The overall construction pipeline of \name.}
    \label{fig:main}
\end{figure*}

\section{\name}
In this section, we introduce our novel benchmark, {\name}, to better evaluate the performance of LLM-based agents under real-world user behaviors. 
{\name} consists of $1638$ high-quality test samples, spanning $59$ executable tool-use environments in multiple domains.
% Each test sample of {\name} is composed of a triplet $(e,t,d)$, denoting environment, task, and user dialogue, respectively. 
Detailed statistics of {\name}, including  domain coverage and distribution of difficulty are deferred to Appendix~\ref{app:statistics}.

In the following, we introduce our detailed procedure to build {\name} from scratch. 
As illustrated in the figure~\ref{fig:main}, the overall construction pipeline consists of three stages: 
(i) \emph{Executable environment and task generation} (Section~\ref{sec:env_task_gen}): We construct standalone, stateful and executable environments based on real-world user queries, and formalize tasks equipped with strict ground-truth annotations.
% such as oracle tool traces with the minimal sequence of required tool invocations, and state transitions that record the expected modification of environment states upon task completion. 
(ii) \emph{Behavior-controlled user dialogue generation}(Section~\ref{sec:user_dial_gen}): Building upon the executable tasks, we synthesize ideal multi-turn user trajectories, which are subsequently perturbed into non-ideal variants guided by the empirical behavior taxonomy defined in Section~\ref{sec: user_taxonomy}. 
(iii) \emph{Iterative generation with multi-level verification} (Section~\ref{sec:filtering}): All generated task pairs undergo a rigorous validation and consistency checking process to guarantee that they are reliably executable, perfectly aligned in underlying task objectives, and free of data leakage.

\subsection{Environment and Task Construction}
\label{sec:env_task_gen}
Our pipeline to construct realistic environments and tasks comprises four stages: query collection, environment construction and initialization, as well as task generation.

\textbf{Query Collection.} 
To ensure our collected queries closely align with real-world scenarios, we curate authentic user queries from three well-established tool-usage datasets: API-Bank \citep{li2023apibankcomprehensivebenchmarktoolaugmented}, ToolAce \citep{liu2025toolacewinningpointsllm}, and Dolci \citep{olmo2026olmo3}.
%Real-world tool use typically relies on a persistent environment that the dynamic data configuration of a system that must be maintained, tracked, and updated across multiple interaction turns. 
To filter out simple queries that do not require interaction with the external environment (e.g., factual QA, single-step text transformation), we developed a statefulness filter, which evaluates whether the successful execution of each query requires querying or modifying the state of the external environment. By applying such filter on all collected queries, we get a clean stateful query set. Further details can be found in the Appendix~\ref{app: Query Collection Prompt Templates}.

% To remove overly simple queries, which require no interaction with external environment, we build a \emph{statefulness filter} $\mathcal{F}_{s}$, which evaluates whether the successful execution of each query requires querying or modifying the states of external environment.
%In this context, state refers to the dynamic data configuration of a system.
%Specifically, each query $q$ is assessed for state dependency: whether successful execution requires querying or modifying persistent environment state across multiple interaction steps. 
%Only queries requiring a persistent, domain-specific environment (e.g., database management, multi-step API orchestration) are retained in the candidate set $\mathcal{Q}$, while 
% By applying $\mathcal{F}_{s}$ on all collected queries, we get a clean query set $\mathcal{Q}$ without stateless queries (e.g., factoid question-answering, single-step text transformation). 
% Details of query selection are left in Appendix~\ref{app: Query Collection Prompt Templates}.

\textbf{Environment Construction.} For each query in the stateful query set, we design a three-step process to synthesize an executable environment: 
(i) \emph{Environment Synthesis Specification}: We prompt the LLM to generate a formal specification comprising state space, toolset, and conditions. Here, the statespace is represented as a series of JSON strings describing entity states, such as \texttt{\{"name": "Dr. Lena", "phone": "555-0142"\}}, the toolset includes descriptions of functionalities and parameters for each callable tool, and the conditions define the preconditions required to invoke these tools.
(ii) \emph{Code Implementation}: We instruct the LLM to compile the specification into an executable Python class with standardized interfaces, enabling seamless integration with the agent framework.
(iii) \emph {Static Validation}: Every synthesized Python class undergoes Abstract Syntax Tree (AST) analysis to verify its syntactic correctness, type consistency, and unified return structure. Environments that pass this verification are added to a pool of verified environments.
Prompt templates and validation protocols are provided in Appendix~\ref{app: Environment Construction Prompt Templates}.

\textbf{State Initialization.} For each verified environment $e$, we generate a diverse bank of initial states $s_0$. This not only yields a collection of diverse evaluation setups but also allows us to directly control the difficulty level of each setup (\text{easy}, \text{medium}, \text{hard}) by varying the number of entities within each state. Every initial state undergoes both direct execution and LLM-based verification to ensure: (i) the environment initializes without error; and (ii) the state contains sufficient structured information to support multi-step tool calls. Prompt details are provided in the Appendix~\ref{app: state initialzation prompt templates}.

\textbf{Task Generation.} For each environment $e$ and its corresponding state initializations $s_0$, we generate a diverse set of tasks $t$, which describe specific objectives that the user expect the agent to accomplish. To guarantee diversity and enable fine-grained difficulty control, we utilize an LLM to generate task descriptions with different numbers of minimum tool-calls. 

To facilitate verification, tasks are subsequently annotated with the following five dimensions using LLM: 
1) \emph{oracle tool trace:} $\tau^* = (a_1^*, \ldots, a_K^*)$ where each tool action $a^*_i \in \mathcal{O}_e$, donating the minimal sequence of tool invocations required to accomplish the task; 
2) \emph{optimal action budget:} an optimal action budget $n_t=\|\tau^*\|$ represent the minimum number of tool invocation; 
3) \emph{state delta:} $\Delta s = s_K \setminus s_0$, which records the difference between the final state $s_K$ and the initial state $s_0$, including the creations, deletions and updates of entities; 
4) \emph{outcome assertions:} $\mathcal{A}_t = \{(\omega_i, v_i)\}$ list specific conditions that must hold true upon task completion, mapping critical state entities $\omega_i$ to their expected values $v_i$. For example, an assertion can be defined as \texttt{(Room\_101, "Booked")}; and 
5) \emph{action constraint graph:} $G_t = (V_t, E_t)$, which is a directed graph that enforces logical compliance throughout the interaction. 
The node set $V_t$ comprises all available tools within the environment, and the directed edges $E_t$ encode strict precedence constraints (e.g., checking room availability must precede confirming the booking) or mutually exclusive action relationships (e.g., approving and rejecting a request within an irreversible system).

The resulting task set establishes a standardized, executable foundation for downstream dialogue generation and robustness evaluation.

\subsection{User Dialogue Generation}
\label{sec:user_dial_gen}
% To enable controlled analysis of how ideal and non-ideal user behaviors affect agent performance, we construct user utterances under seven distinct behavioral profiles in Section~\ref{sec: user_taxonomy}. 
% To do so, we first ask an LLM to simulate a rational user based on the task description $d$ and oracle trace $\tau^*$, then we rewrite user utterance to simulate six non-ideal user behaviors.
% Concretely, we first ask an LLM to simulate an ideal user based on the task description and oracle trace, generating their utterance, then we rewrite user utterance to simulate six non-ideal user dialogues. These process is detailed as:

To analysis the agent performance under ideal and non-ideal user behaviors in real-world scenarios, we construct user dialogues $d$ based on the seven distinct behavioral profiles mentioned in Section~\ref{sec: user_taxonomy}. An example of the resulting dialogues is presented in Figure~\ref{fig:case}. The overall process consists of the following two steps:

%we first ask an LLM to expand task $t$ and oracle trace $\tau^*$ 
%To translate formal tasks into natural language interactions, we first employ a user simulator LLM to expand each task $t \in \mathcal{T}$ into a baseline stable dialogue that strictly follows the oracle trace $\tau^*$ and clearly expresses the underlying intent. 
%Then we derive six controlled unstable behavior variants via behavior-specific rewriting based on the stable dialogue, corresponding to underspecification, information overload, fabricated parameters, goal switching, contradictory constraints, and impatience/hostility. An example of the resulting dialogues is shown in Figure~\ref{fig:case}.

\begin{itemize}[leftmargin=1.5em, itemsep=0.1em, topsep=0.1em]
    \item \textbf{Ideal User dialogue Generation}: 
    % Firstly, we segment the standard tool invocation sequence to derive intermediate goals for each step. Then, we prompt an LLM to generate the corresponding ideal utterances. These utterances explicitly convey the intermediate goals and naturally encapsulate the underlying execution logic, while strictly avoiding any leakage of tool names, internal identifiers, or system state information. In this way, we can produce a complete set of ideal user utterances for each task.
    To generate $T$-turn ($T \in \mathbb{N}^{+}$) ideal user dialogues for a specific task $t$, we first partition the oracle tool trace into $T$ non-overlapping tool segments. We then prompt an LLM to generate an utterance for each segment that encodes the execution logic of the underlying tool segment, while strictly avoiding any leakage of tool names, internal identifiers, or system state information, yielding $T$ user utterances in total.

    \item \textbf{Non-ideal User Dialogue Generation}: 
    % Firstly, we sample from a set of non-ideal behaviors and then prompt an LLM to inject these behaviors into the ideal utterances, thereby producing non-ideal user utterances. The injection strategy for each of the six non-ideal user behaviors is detailed in Appendix~\ref{app: Injection Strategy for Non-ideal User Behavior}.
    Each non-ideal user dialogue is derived from an ideal user dialogue by applying controlled rewriting to its utterances. The detailed rewriting strategy for the six types of non-ideal user behaviors are provided in Appendix~\ref{app: Injection Strategy for Non-ideal User Behavior}.
    
\end{itemize}

\subsection{Verification and Iterative Modification}
\label{sec:filtering}

% For each set of environment $e$, task $t$, and dialogue pairs $(u^+, u^-)$, we perform two final checks before adding them to {\name}.
For each set of environment, task, and dialogues, we perform the following three-stage verifications before adding them to {\name}.

\begin{itemize}[leftmargin=1.5em, itemsep=0.1em]
    \item \textbf{Environment Stability Verification}: We prompt an LLM to exhaustively attempt all available tools within the environment under multiple randomly sampled initial states. Subsequently, we introduce a white-box evaluator LLM, which has full access to the underlying environment and task information, to verify the correctness of both the tool outputs and the resulting state transitions. We iteratively refine the environment specification until they fully pass this verification.
    \item \textbf{Task Consistency and Achievability Verification}: We execute the oracle tool trace to verify that it successfully leads to the expected state changes and satisfies all outcome assertions under the constraints of the environment conditions and action constraint graph. If a task fails this verification, we iteratively revise the task blueprint until it fully meets the requirements.
    \item \textbf{User Dialogue Logical Consistency and Validity Verification}: We further verify the logical consistency and validity of each user dialogue using a white-box LLM-based evaluator, which performs a rigorous semantic and structural check. It enforces that: (1) the underlying intents strictly align with the task description, and the $T$-turn utterances are logically sequenced, non-overlapping, and collectively sufficient to achieve the final objective; (2) all user intent and tool invocation parameters remain inferable from the current turn or preceding context; and (3) no internal tool names, identifiers, or backend schemas are leaked. Any dialogue violating these criteria is regenerated.
\end{itemize}

Finally, we conduct multi-level consistency filtering to further improve the sample quality. Additional implementation details and prompt templates are provided in Appendix~\ref{app: Multi-Level Consistency Filtering} and Appendix~\ref{app: Prompt Templates for Dual-Task Filtering}.

\section{Evaluation with \name}
%\subsection{Evaluation Settings}

%\noindent\textbf{Benchmark Details.}
%和jWe constructed 59 executable environments and 234 base tasks, each instantiated into one stable dialogue $d^{+}$ alongside six unstable variants $\{d^-_1, \dots, d^-_6\}$ representing different unstable behaviors, yielding $1638$ task instances that span 11 task domains. Key statistical observations are as follows: \textit{i)} Dialogue modes of stable and unstable variants are equal, with difficulty stratified into easy (25.8\%), medium (49.2\%), and hard (25.0\%), ensuring diversity and challenge across the benchmark. \textit{ii)} The 11 task domains cover healthcare, finance, enterprise/CRM, infrastructure/DevOps, media streaming, social platforms, e-commerce, gaming, sports, transportation, and travel/hospitality, all corresponding to commonly encountered real-world stateful tool-use scenarios. \textit{iii)} The average user-turn count is 1.69 and the average oracle action turns $n_t$ is 5.97, with 3.36 tool-call steps per task, reflecting the genuinely multi-step, state-modifying nature of the underlying interactions.

In this section, we give a comprehensive guide on the usage of {\name} to evaluate LLMs' performance when dealing with non-ideal users. 

\subsection{Evaluation Procedure}

During evaluation, the agent receives a system prompt detailing the environment, constraints, and available toolset. Without direct access to the initial state $s_0$, the agent must interactively infer necessary context. At each step, based on the accumulated message history, it either invokes an API or replies to the user in natural language. This iterative process continues until all user queries are addressed or the step limit is reached. Upon termination, the terminal state, state delta, and tool trace are logged for evaluation.

\begin{figure*}[!t]
    \centering
    \includegraphics[width=\textwidth]{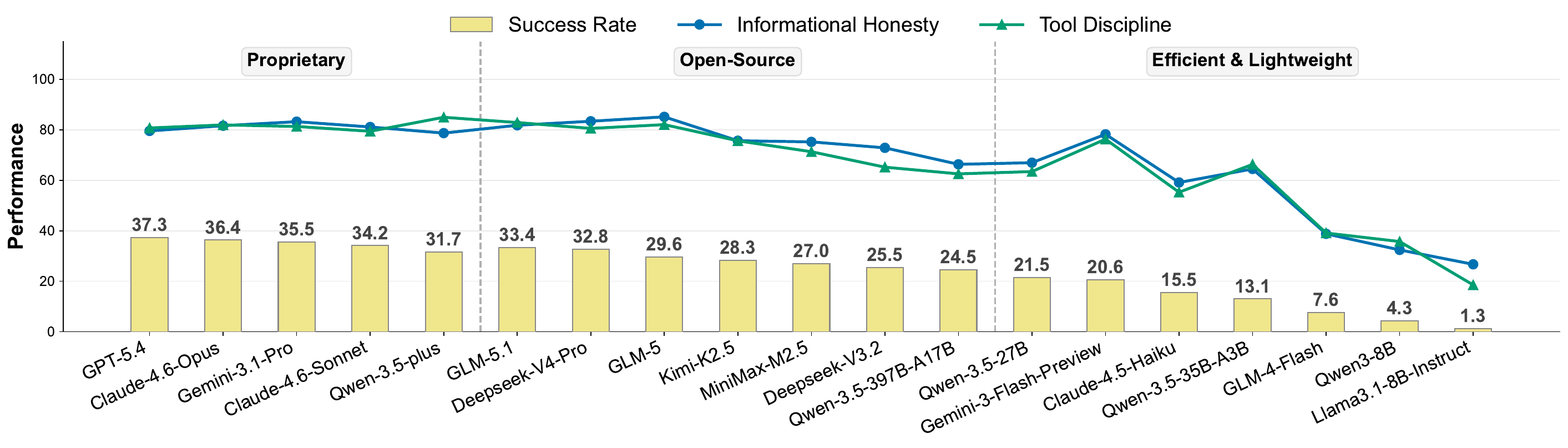}
    \caption{Success Rate, Informational Honesty, and Tool Discipline of the 19 evaluated models on {\name}.}
    \label{fig:main_exp_fig}
\end{figure*}

\subsection{Evaluation Metrics}
For a task and user dialogue, the agent generates a tool trace $\hat{\tau}$ and a final state difference $\Delta \hat{s}$. The result-oriented overall reward $r$ is based on the success rate, which is a binary indicator measuring task completion. An execution is marked successful if and only if it satisfies the following criteria: 
(i) \textit{Action Coverage}: the predicted trace $\hat{\tau}$ contains all essential tool actions specified in $\tau^*$; 
(ii) \textit{Logical Compliance}: the order of tool calls in $\hat{\tau}$ strictly conforms to the dependency edges in the constraint graph $E_t$; and 
(iii) \textit{State Verification}: the final environment state difference $\Delta \hat{s}$ fulfills all outcome assertions defined in $\mathcal{A}_t$. 
% (2) Turn Efficiency $r_{te}$, which measures the interaction cost to evaluate how efficiently the agent reaches the goal. We define $\text{TE} = \min(1, \frac{n_t}{\hat{n}})$, where $n_t$ is the optimal action budget, and $\hat{n}$ is the actual action budget of agent tool calls.
% \end{itemize}

% \noindent\textbf{(1) Effectiveness.} It measures whether the agent correctly and efficiently accomplishes the task. It is composed of two quantitative metrics: 

% \begin{itemize}[leftmargin=10pt]
%     \item \textbf{Success Rate (SR)} is a binary indicator ($\text{SR} \in \{0, 1\}$) measuring task completion. An execution is marked successful if and only if it satisfies the following criteria: (i) \textit{Action Coverage}: the predicted trace $\hat{\tau}$ contains all essential tool actions specified in $\tau^*$; (ii) \textit{Logical Compliance}: the order of tool calls in $\hat{\tau}$ strictly conforms to the dependency edges in the constraint graph $E_t$; and (iii) \textit{State Verification}: the final environment state difference $\Delta \hat{s}$ fulfills all outcome assertions defined in $\mathcal{A}_t$.

%     \item \textbf{Turn Efficiency (TE)} measures the interaction cost to evaluate how efficiently the agent reaches the goal. We define $\text{TE} = \min(1, \frac{n_t}{\hat{n}})$, where $n_t$ is the optimal action budget, and $\hat{n}$ is the actual action budget of agent tool calls.
% \end{itemize}

Besides success rate, we add two additional diagnostic metrics to evaluate response reliability and user experiences, including: 
(1) Informational Honesty $r_{ih}\in(0,1)$, which evaluates whether the agent's responses are strictly grounded in the given context and consistent across dialogue turns. The model is penalized for generating unsupported facts, such as hallucinating non-existent parameters or fabricating system capabilities. 
(2) Tool Discipline $r_{td}\in(0,1)$, which penalizing blind decisions, unauthorized operations, or breaking tool constraints. Our diagnostic metrics are evaluated by LLM-as-a-Judge. The detailed prompt templates can be found in Appendix~\ref{app: Prompt Templates for Reliability Judge}.

\section{Experiments and Analysis}

% We comprehensively evaluate existing agent ability in realistic interactive scenarios, 
We evaluate 19 mainstream open-source and closed-source LLMs, including Claude-4.6-Opus~\citep{anthropic2026opus46}, GPT-5.4~\citep{openai2026gpt54}, and Deepseek-V4-Pro~\citep{deepseek2026v4}. The detailed settings are deferred to Appendix~\ref{app:exp_setting}.

\subsection{Main Results}
\label{sec:main_results}
As shown in Figure \ref{fig:main_exp_fig}, \name\ reveals critical gaps in real-world user experience, which can be summarized into the following key observations:

\emph{(i) Current models are struggled to handle the behaviors of non-ideal users in real-world.} Even the best-performing model, GPT-5.4, achieves a score of only $37.3\%$, indicating a massive margin for future improvement. While the open-source models generally underperform compared to proprietary ones, flagship open-source models such as GLM-5.1 and DeepSeek-V4-Pro exhibit remarkable competitiveness. Furthermore, performance are highly sensitive to model scale: medium-sized models trail frontier models by approximately 15 points, with degradation becoming even more severe in lightweight models. Interestingly, small dense models perform on par with large sparse models (e.g., Qwen-3.5-27B matching Qwen-3.5-397B-A17B), likely due to the latter's lower number of activated parameters. Collectively, these disparities suggest that current LLMs still struggle to maintain robustness and task completion when faced with non-ideal user behaviors.

\emph{(ii) Faithfulness to dialogue contexts and tool constraints is not sufficient.} 
Despite high scores in information honesty and tool discipline, frontier models exhibit relatively low success rates. The primary bottleneck is a failure to adhere to strict procedural ordering. Models often display execution overconfidence by invoking state-modifying tools without prerequisite lookups, violating the read-before-write precedence. Conversely, they exhibit execution paralysis when facing ambiguous constraints. Likely due to safety alignment penalties, models default to refusing requests rather than proactively resolving uncertainty, which severely undermines their autonomous problem-solving capabilities.

\emph{(iii) Lightweight models collapse in both task execution and behavioral reliability.} 
Lightweight dense models score poorly across all metrics, frequently fabricating parameters and violating constraints. Their primary failure mode is an inability to sustain execution trajectories due to constrained reasoning; agents often stall after initial information gathering, leaving tasks partially completed. Additionally, they exhibit \textit{blind state modification} by writing to the environment without prerequisite queries. In contrast, Gemini-3-Flash-Preview achieves high diagnostic scores comparable to top models, demonstrating strong tool discipline and hallucination detection. However, its overall success rate remains low, as failures typically stem from violating invocation order or performance degradation in the final writing phase.

\definecolor{lightgraycol}{RGB}{235,235,235}
\definecolor{darkgraycol}{RGB}{210,210,210}
\definecolor{grouporange}{RGB}{255,229,204}
\definecolor{groupblue}{RGB}{217,234,247}
\definecolor{grouppink}{RGB}{248,221,229}
\definecolor{groupgreen}{RGB}{220,237,220}

\begin{table*}[t!]
\centering
\small
\setlength{\tabcolsep}{4pt}
\renewcommand{\arraystretch}{1.15}
\resizebox{\textwidth}{!}{
\begin{tabular}{
l
ccccccc
>{\columncolor{darkgraycol}}c
}
\toprule
\textbf{Models} 
& \textbf{Ideal} 
& \textbf{Contradict.} 
& \textbf{Goal Switch.} 
& \textbf{Info. Overload} 
& \textbf{Underspec.} 
& \textbf{Impatience} 
& \textbf{Fabricated} 
& \textbf{Overall} \\
\midrule
\rowcolor{grouporange}
\multicolumn{9}{c}{\textit{Proprietary General Models}} \\
GPT-5.4           & 44.02 & \textbf{40.17} {\scriptsize($-$8.7\%)}  & \textbf{41.45} {\scriptsize($-$5.8\%)}  & \textbf{38.46} {\scriptsize($-$12.6\%)} & 24.78 {\scriptsize($-$43.7\%)}          & 37.17 {\scriptsize($-$15.6\%)}          & \textbf{35.47} {\scriptsize($-$19.4\%)} & \textbf{37.30} \\
Claude-Opus-4.6   & \textbf{44.87} & 35.47 {\scriptsize($-$21.0\%)}          & 37.18 {\scriptsize($-$17.1\%)}          & 37.61 {\scriptsize($-$16.2\%)}          & \textbf{28.63} {\scriptsize($-$36.2\%)} & 38.89 {\scriptsize($-$13.3\%)}          & 32.05 {\scriptsize($-$28.6\%)}          & 36.38 \\
Gemini-3.1-Pro    & 44.44 & 32.05 {\scriptsize($-$27.9\%)}          & \textbf{41.45} {\scriptsize($-$6.7\%)}  & 35.89 {\scriptsize($-$19.2\%)}          & 24.78 {\scriptsize($-$44.2\%)}          & \textbf{39.74} {\scriptsize($-$10.6\%)} & 30.34 {\scriptsize($-$31.7\%)}          & 35.53 \\
Claude-Sonnet-4.6 & 41.45 & 27.78 {\scriptsize($-$33.0\%)}          & 40.59 {\scriptsize($-$2.1\%)}           & 36.32 {\scriptsize($-$12.4\%)}          & 21.79 {\scriptsize($-$47.4\%)}          & 38.03 {\scriptsize($-$8.3\%)}           & 33.33 {\scriptsize($-$19.6\%)}          & 34.18 \\
Qwen3.5-Plus      & 39.74 & 32.91 {\scriptsize($-$17.2\%)}          & 35.90 {\scriptsize($-$9.7\%)}           & 31.20 {\scriptsize($-$21.5\%)}          & 17.52 {\scriptsize($-$55.9\%)}          & 36.75 {\scriptsize($-$7.5\%)}           & 27.78 {\scriptsize($-$30.1\%)}          & 31.68 \\
\rowcolor{lightgraycol}
\textit{Avg.}     & 42.90 & 33.68 {\scriptsize($-$21.5\%)}          & 39.31 {\scriptsize($-$8.4\%)}           & 35.90 {\scriptsize($-$16.3\%)}          & 23.50 {\scriptsize($-$45.2\%)}          & 38.12 {\scriptsize($-$11.1\%)}          & 31.79 {\scriptsize($-$25.9\%)}          & 35.01 \\
\midrule
\rowcolor{groupblue}
\multicolumn{9}{c}{\textit{Open-Source General Models}} \\
GLM-5.1             & 41.03 & \textbf{34.19} {\scriptsize($-$16.7\%)} & 34.19 {\scriptsize($-$16.7\%)}          & 32.48 {\scriptsize($-$20.8\%)}          & 21.79 {\scriptsize($-$46.9\%)}          & 36.32 {\scriptsize($-$11.5\%)}          & \textbf{33.76} {\scriptsize($-$17.7\%)} & \textbf{33.39} \\
DeepSeek-V4-Pro     & \textbf{41.88} & 32.47 {\scriptsize($-$22.5\%)}          & \textbf{37.60} {\scriptsize($-$10.2\%)} & \textbf{33.76} {\scriptsize($-$19.4\%)} & \textbf{23.51} {\scriptsize($-$43.9\%)} & \textbf{40.59} {\scriptsize($-$3.1\%)}  & 19.65 {\scriptsize($-$53.1\%)}          & 32.78 \\
GLM-5               & 36.75 & 26.92 {\scriptsize($-$26.7\%)}          & 31.62 {\scriptsize($-$14.0\%)}          & 32.91 {\scriptsize($-$10.4\%)}          & 19.23 {\scriptsize($-$47.7\%)}          & 33.76 {\scriptsize($-$8.1\%)}           & 26.06 {\scriptsize($-$29.1\%)}          & 29.60 \\
Kimi-K2.5            & 37.18 & 29.49 {\scriptsize($-$20.7\%)}          & 30.34 {\scriptsize($-$18.4\%)}          & 26.50 {\scriptsize($-$28.7\%)}          & 17.09 {\scriptsize($-$54.0\%)}          & 29.49 {\scriptsize($-$20.7\%)}          & 27.78 {\scriptsize($-$25.3\%)}          & 28.27 \\
MiniMax-M2.5         & 32.91 & 28.63 {\scriptsize($-$13.0\%)}          & 29.06 {\scriptsize($-$11.7\%)}          & 25.21 {\scriptsize($-$23.4\%)}          & 15.38 {\scriptsize($-$53.3\%)}          & 28.63 {\scriptsize($-$13.0\%)}          & 29.06 {\scriptsize($-$11.7\%)}          & 26.99 \\
DeepSeek-V3.2       & 31.62 & 26.07 {\scriptsize($-$17.5\%)}          & 28.21 {\scriptsize($-$10.8\%)}          & 25.21 {\scriptsize($-$20.3\%)}          & 14.10 {\scriptsize($-$55.4\%)}          & 26.92 {\scriptsize($-$14.9\%)}          & 26.07 {\scriptsize($-$17.5\%)}          & 25.46 \\
Qwen3.5-397B-A17B   & 33.33 & 26.07 {\scriptsize($-$21.8\%)}          & 26.07 {\scriptsize($-$21.8\%)}          & 23.93 {\scriptsize($-$28.2\%)}          & 13.68 {\scriptsize($-$59.0\%)}          & 25.64 {\scriptsize($-$23.1\%)}          & 23.08 {\scriptsize($-$30.7\%)}          & 24.54 \\
\rowcolor{lightgraycol}
\textit{Avg.}       & 36.39 & 29.12 {\scriptsize($-$20.0\%)}          & 31.01 {\scriptsize($-$14.8\%)}          & 28.57 {\scriptsize($-$21.5\%)}          & 17.83 {\scriptsize($-$51.0\%)}          & 31.62 {\scriptsize($-$13.1\%)}          & 26.49 {\scriptsize($-$27.2\%)}          & 28.72 \\
\midrule
\rowcolor{groupgreen}
\multicolumn{9}{c}{\textit{Efficient \& Lightweight Models}} \\
Qwen3.5-27B             & \textbf{30.34} & \textbf{20.51} {\scriptsize($-$32.4\%)} & \textbf{23.50} {\scriptsize($-$22.5\%)} & \textbf{22.22} {\scriptsize($-$26.8\%)} & \textbf{16.24} {\scriptsize($-$46.5\%)} & \textbf{21.79} {\scriptsize($-$28.2\%)} & 15.81 {\scriptsize($-$47.9\%)}          & \textbf{21.49} \\
Gemini-3-Flash-Preview  & \textbf{30.34} & 18.80 {\scriptsize($-$38.0\%)}          & 22.22 {\scriptsize($-$26.8\%)}          & 19.66 {\scriptsize($-$35.2\%)}          & 11.97 {\scriptsize($-$60.6\%)}          & \textbf{21.79} {\scriptsize($-$28.2\%)} & \textbf{19.23} {\scriptsize($-$36.6\%)} & 20.58 \\
Claude-4.5-Haiku        & 20.94 & 15.38 {\scriptsize($-$26.6\%)}          & 15.81 {\scriptsize($-$24.5\%)}          & 14.96 {\scriptsize($-$28.6\%)}          & 8.55  {\scriptsize($-$59.2\%)}          & 15.81 {\scriptsize($-$24.5\%)}          & 17.09 {\scriptsize($-$18.4\%)}          & 15.51 \\
Qwen3.5-35B-A3B         & 15.38 & 13.24 {\scriptsize($-$13.9\%)}          & 14.10 {\scriptsize($-$8.3\%)}           & 14.10 {\scriptsize($-$8.3\%)}           & 6.83  {\scriptsize($-$55.6\%)}          & 14.96 {\scriptsize($-$2.7\%)}           & 12.82 {\scriptsize($-$16.6\%)}          & 13.06 \\
GLM-4-Flash             & 9.83  & 7.26  {\scriptsize($-$26.1\%)}          & 8.12  {\scriptsize($-$17.4\%)}          & 7.26  {\scriptsize($-$26.1\%)}          & 4.27  {\scriptsize($-$56.6\%)}          & 8.12  {\scriptsize($-$17.4\%)}          & 8.55  {\scriptsize($-$13.0\%)}          & 7.63  \\
Qwen3-8B                & 6.84  & 4.27  {\scriptsize($-$37.6\%)}          & 3.42  {\scriptsize($-$50.0\%)}          & 4.70  {\scriptsize($-$31.3\%)}          & 2.56  {\scriptsize($-$62.6\%)}          & 4.27  {\scriptsize($-$37.6\%)}          & 4.27  {\scriptsize($-$37.6\%)}          & 4.34  \\
Llama3.1-8B-Instruct    & 1.28  & 1.71  {\scriptsize($+$33.6\%)}          & 1.28  {\scriptsize($-$0.0\%)}           & 1.28  {\scriptsize($-$0.0\%)}           & 0.43  {\scriptsize($-$66.4\%)}          & 1.71  {\scriptsize($+$33.6\%)}          & 1.28  {\scriptsize($-$0.0\%)}           & 1.28  \\
\rowcolor{lightgraycol}
\textit{Avg.}           & 16.42 & 11.60 {\scriptsize($-$29.4\%)}          & 12.64 {\scriptsize($-$23.0\%)}          & 12.03 {\scriptsize($-$26.7\%)}          & 7.26  {\scriptsize($-$55.8\%)}          & 12.64 {\scriptsize($-$23.0\%)}          & 11.29 {\scriptsize($-$31.2\%)}          & 11.98 \\
\bottomrule
\end{tabular}
}
\caption{Success Rate of 19 models across 7 user-behavior categories. 
% \textbf{Overall} (highlighted) is the aggregated score across all behaviors.
% \textbf{Bold} indicates the best result within each model group. \textit{Avg.} rows report the mean score and relative drop within each group; 
Values in parentheses indicate the relative performance drop with respect to the \textbf{Ideal} baseline.
% \textbf{Behaviors:} Ideal, Contradictory Constraints, Goal Switching, Information Overload, Underspecification, Impatience \& Hostility, Fabricated Parameters.
}
\label{tab:behavior_pass1}
\end{table*}

\subsection{Impact of Non-ideal User Behavior}

We comprehensively analyze the impact of non-ideal user behaviors on function-calling capabilities. As shown in Table~\ref{tab:behavior_pass1}, different models exhibit distinct robustness and capability divergence when handling fine-grained behavioral perturbations.

\emph{(i) The shift from ideal to non-ideal user behaviors causes a universal performance decline, yet frontier models demonstrate stronger resilience.}
While all evaluated models experience performance degradation when exposed to non-ideal user behaviors, the magnitude of this decline is highly correlates with the model's overall capability. Proprietary models demonstrate a stronger resilience, experiencing the smallest relative performance drops. Leading open-source models follow closely but expose slightly deeper vulnerabilities. In contrast, efficient and lightweight models suffer severe performance collapses, trailing the frontier models heavily. This universal degradation underscores that robust foundational reasoning is the key to resisting complex behavioral noise.
% For instance, GPT-5.4 shows a marginal drop in performance, falling from 44.44 under ideal conditions to 42.02 with non-ideal users. In contrast, the medium-sized Qwen3.5-397B-A17B shows a noticeable decline, falling from 31.62 to 27.92. This analysis demonstrates that larger, top-performing models, by virtue of their superior internal mechanisms, can effectively handle behavioral noise and accurately rectify user intention.

% As shown in Figure~\ref{fig:combined_figure}(a), all evaluated models exhibit some degree of performance degradation when faced with unstable user behaviors. However, the magnitude of this decline is highly dependent on the model's overall capability. For instance, GPT-5.4 scores 44.44 under stable conditions and 42.02 under unstable conditions, a marginal drop of just 2.4. Similarly, GLM-5.1 shows only a slight decrease from 41.45 to 39.24.
% In contrast, the medium-scale Qwen3.5-397B-A17B demonstrates a more pronounced regression, falling from 31.62 to 27.92, a decline of nearly 3.7. Although DeepSeek-V3.2 exhibits a relatively smaller drop, this is primarily attributed to its inherently low baseline performance under stable conditions, which objectively limits the room for further decline.
% These findings indicate that while instability in user behavior imposes a shock on all models, those with larger parameter scales and superior performance can effectively absorb behavioral noise and accurately rectify user intent, thanks to their more robust intrinsic mechanisms.

\emph{(ii) Frontier models exhabit distinct, localized vulnerabilities.} 
Fine-grained behavioral analysis reveals that different models exhibit unique blind spots to specific types of non-ideal behaviors. For instance, the Claude series is particularly fragile against Contradictory Constraints, whereas the Gemini series is more vulnerable on Fabricated Parameters. These diverse, model-specific vulnerabilities highlighting the indispensable role of fine-grained evaluation dimensions in \name for real-world agent deployments.
%DeepSeek-V4-Pro demonstrates an extreme dichotomy: it exhibits the strongest resilience against Impatience and Hostility, but suffers a catastrophic plunge when encountering Fabricated Parameters. 

\emph{(iii) Underspecification is the most challenging behavior, while models exhibit stronger resilience to Goal Switching and Impatience.} Underspecification triggers the most severe degradation across all model tiers. Proprietary models suffer a 45.2\% relative drop, while lightweight models plummet by nearly 60\%. This indicates that current LLMs still struggle to proactively infer missing arguments from preceding contexts, relying heavily on explicit user clarifications. Conversely, Goal Switching and Impatience \& Hostility are less disruptive. 

\subsection{Error Analyses}
\begin{figure}[t]
    \centering
    \includegraphics[width=0.9\linewidth]{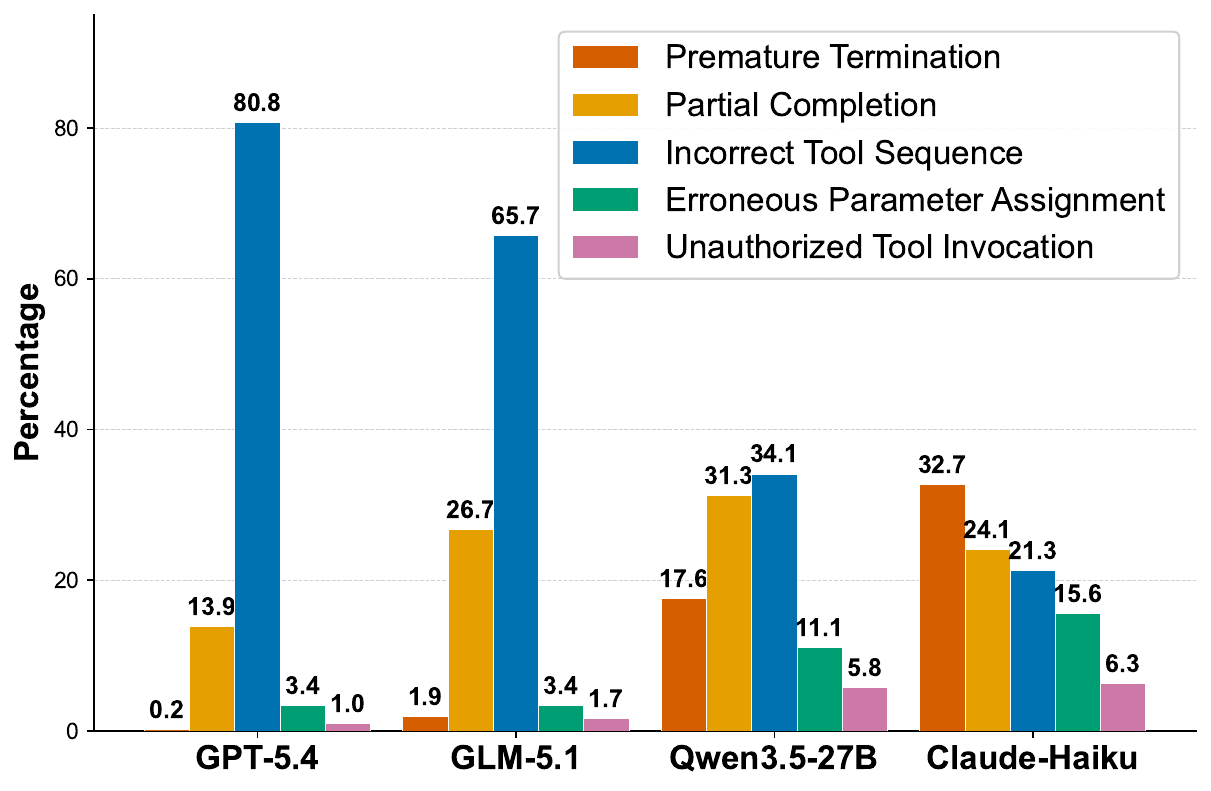}
    \caption{Failure analysis on representative models} 
    \label{fig:failure_cause_analysis}
\end{figure}

To gain deeper insights into these execution bottlenecks, we prompted an LLM (Prompt can be found in Appendix~\ref{app: System Prompt for Error Analyses}) to analyze every failed trajectory from four representative models. As illustrated in Figure~\ref{fig:failure_cause_analysis}, our analysis reveals that these failures fall into five distinct categories. (i) Incorrect Tool Sequence emerges as the universally dominant failure mode, indicating that while models generally comprehend the required semantic actions, they exhibit overconfidence in invoking state-modifying tools and frequently violate strict procedural dependencies. (ii) Premature Termination and Partial Completion constitute another significant cluster of failures. When confronting user uncertainty, models are over-cautious, prematurely halting execution and relying heavily on explicit user clarifications rather than leveraging proactive self-inference to propel the task forward. For weaker models, another reason lies in their inability to sustain reasoning momentum over extended contexts. (iii) Erroneous Parameter Assignment and Unauthorized Tool Invocation predominantly afflict lightweight models. Rather than extracting precise environment identifiers, they tend to inject raw natural language directly into API arguments or hallucinate generic placeholder IDs. The primary reason lies in their lack of multi-step reasoning capabilities, which drives them to adopt a "shortcut" strategy.

\section{Conclusion}
We present {\name}, a benchmark designed to evaluate tool-using agents under realistic non-ideal user interactions spanning seven user-behavior categories. Comprehensive evaluation of 19 mainstream LLMs demonstrates that non-ideal user behaviors pose a fundamental challenge, with all models exhibiting substantial degradation relative to the ideal-user. Through fine-grained error analysis, we identify three failure modes that consistently separate robust models from failing ones. We hope {\name} serves as a foundation for building LLM agents that remain reliable and faithful under the heterogeneous and unpredictable user behaviors encountered in real-world deployment.

\section{Limitations}

While {\name} provides a systematic evaluation of large language model agents under non-ideal interactions, several methodological limitations warrant consideration. Primarily, although the behavioral taxonomy is grounded in authentic interaction logs from the WildChat, {\name} fundamentally relies on LLM-assisted synthesis for generating non-ideal user utterances, which may not fully capture the unpredictable pragmatic nuances and spontaneous disfluencies inherent in genuine human-agent dynamics. Furthermore, the reliance on LLM-as-a-Judge for evaluating diagnostic metrics, such as Informational Honesty and Tool Discipline, introduces potential evaluation biases, prompt sensitivities, and reasoning failures during the assessment of complex multi-turn tool traces. Additionally, the benchmark utilizes deterministic Python class environments that abstract away the operational friction of real-world external APIs, thereby omitting critical deployment variables like network latency, rate limits, and dynamic backend schema shifts. Finally, the current scope is explicitly restricted to English dialogues and seven predefined non-ideal categories, precluding the assessment of agent robustness against malicious adversarial prompt injections or cross-cultural linguistic variations.

\bibliography{custom}

\clearpage
\appendix

\begin{figure*}[!t]
    \centering
    \includegraphics[width= 0.9 \textwidth]{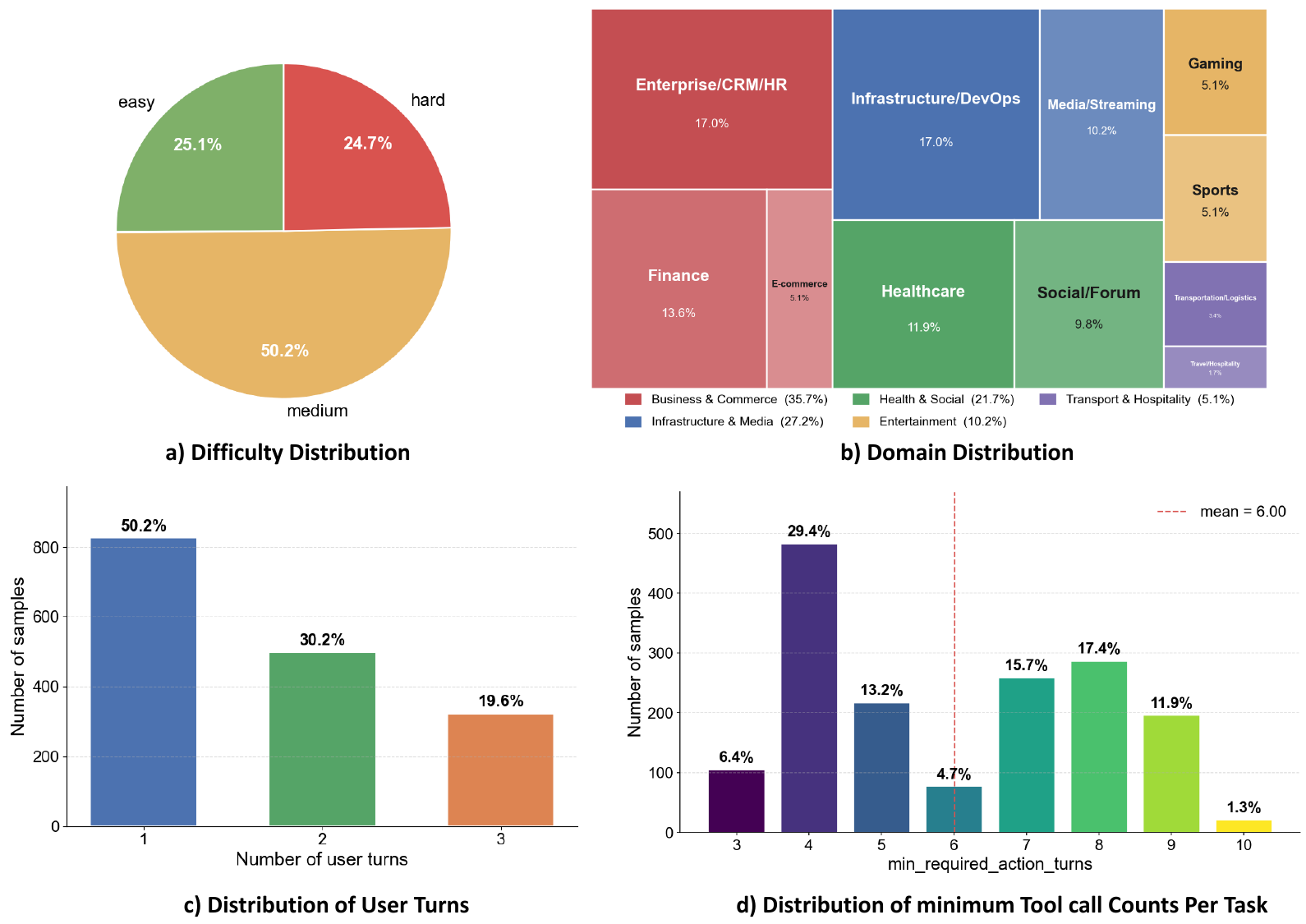}
    \caption{caption.}
    \label{fig:statistics}
\end{figure*}

\section{Statistics of \name}
\label{app:statistics}

\name contains 59 executable environments and 234 base tasks, each instantiated into one ideal rational dialogue alongside six non-ideal variants representing different non-ideal behaviors, yielding $1638$ task instances that span 11 task domains. Key statistical observations are as follows: \textit{i)} Dialogue modes of ideal and non-ideal variants are equal, with difficulty stratified into easy (25.1\%), medium (50.2\%), and hard (24.7\%), ensuring diversity and challenge across the benchmark. \textit{ii)} The 11 task domains cover healthcare, finance, enterprise/CRM, infrastructure/DevOps, media streaming, social platforms, e-commerce, gaming, sports, transportation, and travel/hospitality, all corresponding to commonly encountered real-world stateful tool-use scenarios. \textit{iii)} The average user-turn count is 1.69 and the average optimal action budget $n_t$ is 5.97, reflecting the genuinely multi-step, state-modifying nature of the underlying interactions. The detailed statistics of \name can be found in Figure~\ref{fig:statistics}.

\section{Injection Strategy for Non-ideal User Behavior}
\label{app: Injection Strategy for Non-ideal User Behavior}

\noindent \textbf{Underspecification.} We prompt an LLM to rewrite the ideal user utterance by dropping the essential information slots (e.g., times, locations, or identities) that are explicitly stated, producing a shorter utterance that relies on contextual inference or explicitly clarifying.

\noindent \textbf{Information Overload.} Based on the environment and admissible toolset, we prompt an LLM to generate several extraneous information such as environment background or tangential message and fuse this redundant information with the ideal utterance.

\noindent \textbf{Fabricated Parameters.} We first prompt an LLM to generate nonexistent pseudo-parameters for the current utterance and tool segment, then these fabricated parameters are translated into natural language which can be incorporated into ideal user utterance.

\noindent \textbf{Goal Switching.} We first prompt an LLM to generate several side goals that are unrelated to the ideal user utterance based on the admissible toolset. These side goals are then weaved into ideal user utterance as interruptions or topic drifts.

\noindent \textbf{Contradictory Constraints.} We prompt an LLM to generate a set of conditions that are contradictory or conflicting for the ideal user utterance under environment and toolset. Then these conflicting candidates are incorporated into ideal user utterance.

\noindent \textbf{Impatience and Hostility.} We utilize LLM to directly rewrite the original ideal utterance with impatience, blame, or high pressure tones so that it can be more abrasive and oppressive.

Detailed prompt templates for generating each non-ideal behavior are provided in Appendix~\ref{app: Prompt Templates for User Dialogue Generation}.

\section{Multi-Level Consistency Filtering}
\label{app: Multi-Level Consistency Filtering}
After all candidates are successfully generated and executable, we apply a final filtering process to enforce quality and coverage requirements across three levels. \emph{At the sample level}, each item must include a complete oracle trace $\tau^*$, an optimal action budget $n_t$, an expected state difference $\Delta s$, final-state assertions $\mathcal{A}_t$, and an action constrain graph $G_t$. \emph{At the pair level}, a base task $t$ is admitted into $\mathcal{B}$ only if both its ideal variant $d^+$ and all non-ideal variants $d^-$ pass these checks. Any sample or task pair violating these criteria is rigorously excluded.

\section{Experiment Setting}
\label{app:exp_setting}
All models are uniformly configured with a \texttt{default} context window and employ the \texttt{native function calling} strategy for tool calling and multi-turn dialogue management. Inference is deployed on \texttt{Nvidia H200} GPUs, with an API temperature set to \texttt{0} and a maximum generation of \texttt{20} steps.

\section{Prompt Templates for User Behavior Annotation}
\label{app: user behavior annotation prompt}
Figure~\ref{fig:user behavior annotation prompt} provides the prompt for user behavior annotation introduced in section~\ref{sec: user_taxonomy}.

\begin{figure*}[t]

\centering
\begin{tcolorbox}[
    enhanced,
    width=\linewidth,
    arc=1.5mm,
    boxrule=0.5pt,
    colframe=black,
    colback=gray!3,
    title=\textbf{Prompt Templates for User Behavior Annotation},
    fonttitle=\bfseries,
    colbacktitle=gray!15,
    coltitle=black,
    coltext=black,
]

\small
\ttfamily

You are an expert dialogue analyst for stable versus unstable user behavior in realistic tool-use or task-oriented interactions.
Judge only the user's behavior. Use assistant turns only as context.

Return strict JSON only.

[Task]
Analyze the dialogue below and decide whether the user shows any unstable behavior under the six-category taxonomy.

[Scope]
- Evaluate only user behavior. Use assistant messages only as context.
- A dialogue is stable if the user remains clear, sufficiently specified, consistent, and cooperative enough for normal task progress.
- A dialogue is unstable if at least one of the six behaviors below is clearly present.
- A dialogue may have zero, one, or at most two primary labels. Prefer the most specific label(s).
- If evidence is weak or ambiguous, do not assign the label.
- Do not mark a dialogue as unstable merely because the topic is unusual, sensitive, creative, or unsafe.
- For goal\_switching, contradictory\_constraints, and impatience\_and\_hostility, use cross-turn context when available.

[Six Unstable Behavior Labels]

\{taxonomy\_block\}

[Output JSON]

If unstable behavior is present, return:

\begin{Verbatim}[fontsize=\small]
{{
  "is_unstable": true,
  "behaviors": ["underspecification", "goal_switching"],
  "primary_behavior": "underspecification",
  "behavior_evidence": {{
    "underspecification": {{
      "message_indexes": [1],
      "reason": "The user omits required execution details."
    }}
  }},
  "overall_reason": "Short overall explanation."
}}
\end{Verbatim}

If no unstable behavior is present, return:

\begin{Verbatim}
{{
  "is_unstable": false,
  "behaviors": [],
  "primary_behavior": null,
  "behavior_evidence": {{}},
  "overall_reason": "Short overall explanation."
}}
\end{Verbatim}
[Dialogue Metadata]

id: \{dialogue\_id\}

model: \{model\}

message\_count: \{message\_count\}

user\_turn\_count: \{user\_turn\_count\}

original\_turn\_field: \{original\_turn\_field\

[Dialogue]

\{dialogue\_block\}

\end{tcolorbox}
\caption{Prompt for user behavior annotation}
\label{fig:user behavior annotation prompt}
\end{figure*}

\section{Prompt Templates for Environment and Task Construction}
In this section, we provides the system prompts utilized in environment and task construction stage.
\label{app: Prompt Templates for Environment and Task Construction}

\subsection{Query Collection Prompt Templates}
\label{app: Query Collection Prompt Templates}

Figure~\ref{fig: Prompt templates for query collection} is the prompt we use for \emph{statefulness filter}.

\begin{figure*}[t]

\centering
\begin{tcolorbox}[
    enhanced,
    width=\linewidth,
    arc=1.5mm,
    boxrule=0.5pt,
    colframe=black,
    colback=gray!3,
    title=\textbf{Query Collection},
    fonttitle=\bfseries,
    colbacktitle=gray!15,
    coltitle=black,
    coltext=black,
]

\small
\ttfamily

You are a system that filters natural language tasks to determine if they are **state-dependent, actionable requests** within a **persistent, domain-specific environment**.

Core Definition

We are ONLY looking for tasks that meet **all** of the following criteria:

1. Persistent Environment — The query is about a domain where:

   - There is a live, ongoing state that can be read or changed
   
   - The environment supports both:
   
     a) Information queries about current state (read operations)  
     b) Explicit state-changing actions (create, update, delete, move, cancel, etc.)

2. State Dependency — The task cannot be answered correctly without:

   - Inspecting the actual current data or configuration in the environment, and/or
   
   - Executing an operation that modifies that data.

3. Domain Specificity — The environment is not general-purpose knowledge; it is a structured system such as:

   - File management system with stored files/folders
   
   - Order/logistics tracking system
   
   - Calendar/scheduling system
   
   - CRM, inventory, ticketing, project management tools
   
   - Other specialized platforms with records that persist over time

4. Actionability in Context — The query must correspond to an actionable operation or status check **within the actual environment** (not hypothetical).

Eligible Task Types

- State queries: "Is invoice 1024 paid?" / "What meetings are scheduled for Wednesday?"

- State modification operations: "Upload the proposal.pdf to the project folder" / "Cancel order 4512" / "Move meeting to 3 PM"

Explicit Exclusions

A request is NOT eligible if it is:

- Open-domain factual Q\&A unrelated to a live state ("Who invented AI?", "What’s the capital of France?")

- Casual conversation ("How are you?", "What's the weather?")

- Content creation ("Write me a story", "Make a poem")

- Pure hypothetical without actual environment interaction

- Isolated reasoning or calculations without accessing persisted state

Judgment Rule — Be strict:

Choose YES only if:

- The query cannot be answered from general knowledge alone

- AND it requires real-time access to persistent state in a domain-specific environment

- AND it targets an actionable operation (either a read or a write to that environment)

- AND the environment has the capability for both queries and modifications

If any criterion is missing → NO.

---

Task
Given a query, first analyze whether it implies or requires:

- A domain-specific environment with both query and modification capabilities

- Accessing or updating persistent state

- Performing a concrete, actionable operation

Then give your final judgment.

Output Format (Strictly enforce)

Analysis

<Detailed reasoning whether this query depends on persistent state, involves a stateful operation, and needs a capable environment as defined>

Answer

YES   --> Only if all strict criteria are met  

NO    --> Otherwise

Analyze the following task and determine if it is a task that depends on a stateful and domain-specific environment.

\{query\}

\end{tcolorbox}
\caption{Prompt templates for query collection}
\label{fig: Prompt templates for query collection}
\end{figure*}

\subsection{Environment Construction Prompt Templates}
\label{app: Environment Construction Prompt Templates}

We first infer the environment specification from the stateful query, with the prompt shown in Figure~\ref{fig: Prompt template for inferring environment specification}. Next, based on the inferred environment specification, we derive the attributes of the entities (Figure~\ref{fig: Prompt templates for inferring entity attributes}) and the tool specifications (Figure~\ref{fig: Prompt templates for inferring tool specification}). Finally, we compile the entities and tools into executable Python classes, as shown in Figure~\ref{fig: Prompt templates for inferring entity attribute python code} and Figure~\ref{fig: Prompt templates for inferring python code for tools}.

\begin{figure*}[t]

\centering
\begin{tcolorbox}[
    enhanced,
    width=\linewidth,
    arc=1.5mm,
    boxrule=0.5pt,
    colframe=black,
    colback=gray!3,
    title=\textbf{Infer Environment Specification},
    fonttitle=\bfseries,
    colbacktitle=gray!15,
    coltitle=black,
    coltext=black,
]

\small
\ttfamily

You are a Task Analyst.  
Given a raw task description, your objective is to identify the most plausible stateful and domain-specific environment in which this task would naturally occur.  

The chosen environment should strike a balance: not so broad as to be meaningless, and not so narrow as to apply only to a single, highly specific case. It should be scoped such that this task, along with similar related tasks, can be executed meaningfully.  

Guidelines:  

- If multiple environments seem equally plausible, select one at random rather than listing all possibilities.  

- Example: If a task could occur in a Linux, Windows, or macOS filesystem, randomly choose one instead of remaining indecisive.  

Your response must include the following sections:  

1. Analysis  

   - Explain the reasoning process used to connect the task to the chosen environment. 
   
   - Note any relevant entities, constraints, relationships, or dynamics implied by the task.  

2. Environment Summary  

   - Provide a concise label for the environment type. 
   
   - Examples: Linux filesystem, E-commerce order management system, Airline booking system.  

3. Environment Introduction  

   - Introduce the environment itself, without referring to the current task.  
   
   - Focus on its inherent structure, the nature of the state it maintains, typical operations it supports, and its general scope in real-world usage.  
   
   - Limit to approximately three sentences.  

4. Metrics  

   - Usefulness: 1-10  
   
     Reflects how broadly applicable and valuable this environment is in real-world scenarios. Higher scores indicate environments relevant to many contexts and industries.  
     
   - Modelability: 1-10  
   
     Indicates how straightforward it would be to represent this environment using a single Python class, with attributes holding state and methods performing reading, writing, and querying operations. Higher scores indicate simpler, more self-contained structures.  

Your response must follow exactly this format, with no additional text or commentary:

Analysis

[Your analysis]

Environment Summary

[Your environment summary]

Environment Introduction

[Your environment introduction]

Metrics

Usefulness: [1-10]

Modelability: [1-10]

Analyze the following task and infer the most plausible stateful and domain-specific environment in which this task would naturally take place

\{task\}

\end{tcolorbox}
\caption{Prompt template for inferring environment specification}
\label{fig: Prompt template for inferring environment specification}
\end{figure*}

\begin{figure*}[t]

\centering
\begin{tcolorbox}[
    enhanced,
    width=\linewidth,
    arc=1.5mm,
    boxrule=0.5pt,
    colframe=black,
    colback=gray!3,
    title=\textbf{Infer Entity Attributes},
    fonttitle=\bfseries,
    colbacktitle=gray!15,
    coltitle=black,
    coltext=black,
]

\small
\ttfamily

You are an expert task and environment analyst.  

Given an environment description and a example task in this environment, infer the set of state variables (state space) that the environment maintained.  

The state should not be too broad (e.g. "all possible data in an e-commerce system"), nor too narrow (only for this single task).  Instead, reasonably design it to support this task and similar tasks in the same environment.  

The input format is:

Environment Summary

[Environment summary]

Environment Introduction

[Environment introduction]

A Example Task in This Environment

[Example task]

Your output must follow the format below (do not include any other text):

Analysis

[Your thought process: What states are involved in the environment? What entities/attributes need to be tracked? What constraints or rules exist in the environment? ……]

State Space Definition

- Entity: EntityName1  

  - Attributes: Attribute1, Attribute2, ...
  
  - Description: The role of this entity in the environment

- Entity: EntityName2

  - Attributes: ...
  
  - Description: ...

Constraints \& Rules

- Constraint 1

- Constraint 2

...

Analyze the following task and environment, and infer the set of state variables (state space) that the environment maintained.  

Environment Summary
\{env\_summary\}

Environment Introduction

\{env\_introduction\}

A Example Task in This Environment

\{task\}

\end{tcolorbox}
\caption{Prompt templates for inferring entity attributes}
\label{fig: Prompt templates for inferring entity attributes}
\end{figure*}

\begin{figure*}[t]

\centering
\begin{tcolorbox}[
    enhanced,
    width=\linewidth,
    arc=1.5mm,
    boxrule=0.5pt,
    colframe=black,
    colback=gray!3,
    title=\textbf{Infer Tool Specification},
    fonttitle=\bfseries,
    colbacktitle=gray!15,
    coltitle=black,
    coltext=black,
]

\small
\ttfamily
You are an expert in building and analyzing agent environments.

Given an environment summary, introduction, state space definition, constraint rules, Python base class definition, and example task, your goal is to analyze the current environment and then generate the list of operations needed to support the task in this environment (including information query class and state modification class).

Each operation will be converted into a class function for the Agent to use in subsequent steps.

Key Points:  

- Operations are divided into 2 categories: Information Query Class and State Change Class.  

- Each operation includes: operation name + brief description.

- Before output, you must first write Analysis: explain task logic → which are query operations, which are state change operations → and how constraints are related.  

Input Format:
Based on the following environment specification, produce the operation list.

\begin{verbatim}
{
  "environment_summary": "...",
  "environment_introduction": "...",
  "state_space_definition": [...],
  "constraints_rules": [...],
  "environment_class_definition": "...",
  "environment_example_task": "...",
}

\end{verbatim}

Strictly maintain the following Output Format:

Analysis
[Explain operation requirements + classification logic + how constraints affect + ……]

Operation List

Information Query Class

- Operation: OperationName Description: xxxx  

- Operation: OperationName Description: xxxx 

- ……

State Change Class

- Operation: OperationName Description: xxxx 

- Operation: OperationName Description: xxxx  

- ……

\end{tcolorbox}
\captionof{figure}{Prompt templates for inferring tool specification}
\label{fig: Prompt templates for inferring tool specification}
\end{figure*}

\begin{figure*}[t]

\centering
\begin{tcolorbox}[
    enhanced,
    width=\linewidth,
    arc=1.5mm,
    boxrule=0.5pt,
    colframe=black,
    colback=gray!3,
    title=\textbf{Infer Entity Attribute Python Code},
    fonttitle=\bfseries,
    colbacktitle=gray!15,
    coltitle=black,
    coltext=black,
]

\small
\ttfamily

You are an AI coding assistant.  

Your job is to translate an environment specification into a Python environment class definition.  

The class should simulate the stateful environment structure (without methods yet).  

You should analyze first and then generate code.

You should follow the rules of Analysis and Code to generate the code.

Rules of Analysis

- Determine the environment class name. It should be EnvironmentSummary or an appropriate adaptation (e.g., `LinuxFileSystem`, `EcommerceOrderSystem`).  

- Extract attribute names (comma-separated) from each entity in `state\_space\_definition`.

- If needed, generate a corresponding `TypedDict` using the extracted attributes, with attribute name → key and attribute value type → inferred from the appropriate Python primitive type (e.g., `id`=str, `name`=str, `category`=str, `price`/`size`=float/int, `quantity`=int, `status`=str, `timestamps`=str/float).

- `constraints\_rules` is left as a comment.

Rules of Code

- Generates each `TypedDict` definition if needed.

- Generates the environment class (with only `\_\_init\_\_' and attributes), with attributes of type `Dict[ID, TypedDict]`.

- Add comments mapping each attribute back to the state space entity/attributes.

- Annotates the constraints in the code comments.

- Do not implement any business logic or methods yet.  

The input format is:

Environment Summary

<short label, e.g. Linux filesystem, E-commerce order system>"

Environment Introduction

<paragraph intro>

State Space Definition
\begin{verbatim}
[
    {
      "entity": "EntityName",
      "attributes": "attr1, attr2, ...",
      "description": "short description"
    },
    ...
]
\end{verbatim}

constraints\_rules

constraint 1 ...

constraint 2 ...

Your output must follow the format below (do not include any other text):

Analysis

[Explains how to design Python environment classes based on tasks and state spaces (including class name selection, mapping entities to data structures, which fields are stored as dict/list, and how constraints are expressed through annotations)]

Class Definition

```python
[Python environment class definition]
```

Given the following Environment, State Space, and Constraints, generate a Python environment class definition accordingly.

Environment Summary

\{env\_summary\}

Environment Introduction

\{env\_introduction\}

State Space Definition

\{state\_space\_definition\}

constraints\_rules

\{constraints\_rules\}

\end{tcolorbox}
\caption{Prompt templates for inferring python code of entity attributes}
\label{fig: Prompt templates for inferring entity attribute python code}
\end{figure*}

\begin{figure*}[t]

\centering
\begin{tcolorbox}[
    enhanced,
    width=\linewidth,
    arc=1.5mm,
    boxrule=0.5pt,
    colframe=black,
    colback=gray!3,
    title=\textbf{Infer Tool Python Code},
    fonttitle=\bfseries,
    colbacktitle=gray!15,
    coltitle=black,
    coltext=black,
]

\small
\ttfamily

You are a code generation assistant.

Given an Agent's environment, including the environment's summary and introduction, the environment's state space definition, the environment's constraint rules, key base class definitions, and the list of operations supported by the environment.

Operations include two types: one is information querying of the environment, and the other is state modification of the environment.

Given one of the operations in the operation list (Target Operation),

You must:  

1. In Analysis, reason about: 

- What entities/attributes are involved.  

- Parameters needed.  

- Expected outputs (queries return structured results, state modifications return success messages). 

- Error/edge cases (e.g., invalid input, permission denied).  

- Does it involve environmental constraints or rules. 

2. In Code, implement the Python method:  

- Method name: `def <operation\_name>(self, ...)`.  Note: Cannot be an independent function, but rather a method function within an already implemented environment class.

- Add clear type hints.  

- Add docstring describing inputs, outputs, constraints. 

- Error handling: do not raise exceptions — return a dict like `\{ "success": False, "error": "reason" \}`.  

- For information-query operations, if successful return `\{ "success": True, "data": <result> \}`.  

- For state-modifying operations, if successful return `\{ "success": True, "message": "operation description" \}`. 

In each subsequent round, the input format is:

Environment Summary

<environment\_summary\_here>

Environment Introduction

<environment\_introduction\_here>

State Space Definition

<state\_space\_definition\_here>

Constraints Rules

<constraints\_rules\_here>

Class Definition]

```python

<class\_definition\_here>

```

Operation List

\{operation\_list\}

Target Operation

\{  "operation\_name": "<operation\_name>",
"operation\_description": "<operation\_description>",
"operation\_type": "<query\_or\_state\_change>"

\}

Your output format must be:

Analysis

[Explain reasoning: inputs, outputs, related entities/attributes, constraints logic, success/failure cases]

Code

```python

def <operation\_name>(self, ...):

<docstring explaining inputs, outputs and constraints>

Implementation

```"

\end{tcolorbox}
\caption{Prompt templates for inferring python code for tools}
\label{fig: Prompt templates for inferring python code for tools}
\end{figure*}

\subsection{State Initialization Prompt Templates}
\label{app: state initialzation prompt templates}
Based on the verified environment, we generate initial states with varying difficulty levels using the system prompt shown in Figure~\ref{fig: System prompt for state initialization}.

\begin{figure*}[t]

\centering
\begin{tcolorbox}[
    enhanced,
    width=\linewidth,
    arc=1.5mm,
    boxrule=0.5pt,
    colframe=black,
    colback=gray!3,
    title=\textbf{State Initialization},
    fonttitle=\bfseries,
    colbacktitle=gray!15,
    coltitle=black,
    coltext=black,
]

\small
\ttfamily

You are an expert environment state initializer for an agent tool-calling benchmark.

Return one valid JSON object that can be passed directly to env.env\_init(init\_config=...).

Requirements:

1. Use the exact top-level state container names provided by the user.

2. Populate realistic, internally consistent, cross-referenced entities.

3. Make the state rich enough for multi-step tool use.

4. Use only fictional data.

5. Output JSON only.

State profile definitions:

Each generated state must match one of the following profiles, as specified in the user message.

- sparse: A minimal but functional state. Populate at least two collections, but keep most collections lean (1–2 entries each). Cross-references between entities should still be valid. Suitable for simple, single-step tasks where the agent needs little data to work with.

- balanced: A moderate state with reasonable variety. Populate multiple collections with 3–5 entries each, include realistic cross-references across entity types, and ensure enough data diversity to support several distinct multi-step tasks. This is the default profile for general-purpose evaluation.

- crowded: A dense, richly populated state. Populate all major collections with 5 or more entries each, include complex and varied cross-references, and reflect realistic operational complexity. Suitable for hard, multi-step tasks that require the agent to filter, compare, or navigate large amounts of data.

When generating the init\_config, strictly conform to the profile specified in the user message. Do not mix profiles or exceed the density implied by the target profile.

\end{tcolorbox}
\caption{System prompt for state initialization}
\label{fig: System prompt for state initialization}
\end{figure*}

\subsection{Task Generation Prompt Templates}
\label{app: Task Generation Prompt Templates}

Figure~\ref{fig: System prompt for task generation} presents the system prompt utilized for task generation process.

\begin{figure*}[t]

\centering
\begin{tcolorbox}[
    enhanced,
    width=\linewidth,
    arc=1.5mm,
    boxrule=0.5pt,
    colframe=black,
    colback=gray!3,
    title=\textbf{Task Generation},
    fonttitle=\bfseries,
    colbacktitle=gray!15,
    coltitle=black,
    coltext=black,
]

\small
\ttfamily

You are an expert benchmark task designer for an agent tool-calling environment.

Design exactly one executable task blueprint and one oracle tool plan.

Requirements:

1. Use only entities that actually exist in the provided init\_config.

2. Prefer tasks with at least one objective state change on structured fields.

3. The user-facing request must not mention internal IDs, tool names, code details, or hidden state keys.

4. The tool plan must be directly executable from the provided initial state, and every tool step must return success during replay.

5. Standard execute tasks should use between 1 and 6 tool calls. A rare 0-tool plan is allowed only when task\_type is clarify\_then\_execute or refuse\_or\_scope.

6. Set agent\_action\_budget to an integer no larger than 50.

7. If a later step needs a newly created entity, explicitly choose its ID in the creation step so the plan stays deterministic.

8. Prefer realistic tasks that require discovery before mutation, but prefer concrete verification tools over high-level helper tools when the state hints already expose the relevant candidate IDs.

9. Avoid plans that depend on unavailable equipment types, nonexistent future slots, or helper tools whose success is uncertain from the current state.

10. If dialogue\_mode is multi\_turn, provide 2-3 coherent turn-level tasks.

11. Set task\_type to one of execute, clarify\_then\_execute, or refuse\_or\_scope. Prefer execute unless the prompt explicitly calls for a clarification-heavy or refusal-heavy task.

12. If task\_type is clarify\_then\_execute or refuse\_or\_scope, include the necessary clarification\_requirements or refusal\_requirements and any forbidden\_actions.

13. If the prompt gives an explicit coverage target for this attempt, satisfy it exactly.

14. Output JSON only.

Return an object with this schema:

\begin{verbatim}
{
  "goal_summary": "short benchmark goal",
  "user_request": "natural user-facing request without IDs",
  "dialogue_mode": "single_turn or multi_turn",
  "difficulty_bucket": "easy or medium or hard",
    "task_type": "execute or clarify_then_execute or refuse_or_scope",
  "tool_call_budget": 2,
  "agent_action_budget": 12,
  "expected_phases": ["phase 1", "phase 2"],
  "expected_final_outcome": "objective expected outcome",
  "relevant_entities": ["entity ids or values central to the task"],
  "quality_checks": "why the task is benchmark-worthy",
  "unique_answer_rationale": "why there is one correct answer before conflict",
  "canonical_answer": "exact expected end result",
  "fallback_answer_hint": "backup answer if natural, else none",
  "fallback_inferior_reason": "why fallback is worse, else none",
    "clarification_requirements": ["required clarification or missing fact, else []"],
    "refusal_requirements": ["required refusal or scope boundary, else []"],
    "forbidden_actions": ["tool/action the agent must not do, else []"],
  "multi_turn_tasks": [
        {
            "task_summary": "turn goal",
            "expected_outcome": "turn outcome",
            "task_type": "execute or clarify_then_execute or refuse_or_scope",
            "required_entities": ["optional entity ids or values"]
        }
  ],
  "tool_plan": [
    {
      "tool_name": "exact tool name",
      "arguments": {"arg": "value"},
      "purpose": "why this tool is called"
    }
  ]
}
\end{verbatim}

\end{tcolorbox}
\caption{System prompt for task generation}
\label{fig: System prompt for task generation}
\end{figure*}

\section{Prompt Templates for User Dialogue Generation}
\label{app: Prompt Templates for User Dialogue Generation}
Based on the task, the detailed system prompt used to generate ideal user dialogue is shown in Figure~\ref{fig: System prompt for stable user dialogue generation}. The non-ideal dialogue variants are then generated by rewriting the ideal dialogue, as illustrated in Figure~\ref{fig: System prompt for unstable user dialogue generation}.

\begin{figure*}[t]

\centering
\begin{tcolorbox}[
    enhanced,
    width=\linewidth,
    arc=1.5mm,
    boxrule=0.5pt,
    colframe=black,
    colback=gray!3,
    title=\textbf{Ideal User Dialogue Generation},
    fonttitle=\bfseries,
    colbacktitle=gray!15,
    coltitle=black,
    coltext=black,
]

\small
\ttfamily
You are a stable user simulator for an agent tool-calling benchmark.

Your job is to realize a validated task blueprint into natural stable user dialogue turns.

Rules:

1. Preserve the exact task semantics and do not invent a different task.

2. Keep the user cooperative, realistic, concise, and grounded in the environment context.

3. Do not mention tool names, internal IDs, hidden schema keys, implementation details, or oracle information.

4. If dialogue\_mode is single\_turn, output exactly one user turn.

5. If dialogue\_mode is multi\_turn, output 2-3 user turns that stay in the same task context.

6. Assign every gold-trace tool step to exactly one user turn using contiguous step groups.

7. Cover all tool steps exactly once, in increasing order.

8. The final assistant response should be a short user-facing completion message consistent with the expected outcome.

9. Output JSON only.

Return an object with this schema:
\begin{verbatim}
{
  "user_turns": [
    {
      "turn_id": 1,
      "user_message": "natural collaborative user message",
      "task_summary": "short turn summary",
      "expected_outcome": "expected turn-level outcome",
      "linked_tool_step_indices": [1, 2]
    }
  ],
  "assistant_final_response": "short user-facing completion message"
}
\end{verbatim}

\end{tcolorbox}
\caption{System prompt for Ideal user dialogue generation}
\label{fig: System prompt for stable user dialogue generation}
\end{figure*}

\begin{figure*}[t]

\centering
\begin{tcolorbox}[
    enhanced,
    width=\linewidth,
    arc=1.5mm,
    boxrule=0.5pt,
    colframe=black,
    colback=gray!3,
    title=\textbf{Non-ideal User Dialogue Generation},
    fonttitle=\bfseries,
    colbacktitle=gray!15,
    coltitle=black,
    coltext=black,
]

\small
\ttfamily

You are an unstable user dialogue rewriter for an agent tool-calling benchmark.

Your job is to rewrite stable user turns into unstable user turns for the exact same task.

Rules:

1. Preserve the exact underlying task for the same task\_id.

2. Keep the same number of user turns and the same turn ordering.

3. Rewrite only the user-facing messages. Do not change the task\_summary, expected\_outcome, linked tool-step alignment, tool plan, or final target state.

4. Inject only the requested six-taxonomy unstable behavior bundle as controlled friction on top of the stable baseline.

5. Do not mention tool names, internal IDs, hidden schema keys, or implementation details.

6. Do not replace the original task with a new mandatory task. Any noisy side remark must still leave the original task recoverable directly or through clarification.

7. Do not leak the oracle trace, backend schema, hidden state, or tool implementation details.

8. Output JSON only.

Return an object with this schema:
\begin{verbatim}
{
  "rewritten_user_turns": [
    {
      "turn_id": 1,
            "user_message": "unstable rewrite of the baseline turn"
    }
  ]
}
\end{verbatim}

Unstable behavior taxonomy:

The following six behaviors define the full set of friction types used in this benchmark. The specific behavior bundle to inject for each dialogue is provided in the user message. Use these definitions to understand the intended effect of each behavior and ensure your rewrite is clearly distinguishable from both the stable baseline and from other behaviors.

- underspecification: Remove essential slots or details from the stable turn while keeping the original task recoverable through context or clarification. The rewritten message should be noticeably vaguer or more ambiguous than the baseline without being unintelligible.

- information\_overload: Fuse the stable request with tangential context, redundant details, or nested side information while preserving the core task. The rewritten message should bury the true request inside irrelevant or excessive content, forcing the agent to filter.

- fabricated\_parameters: Add a natural-language reference to unsupported pseudo-parameters, nonexistent options, or unavailable capabilities without changing the true ground truth. The fabricated element must sound plausible but must not correspond to any real tool or state in the environment.

- goal\_switching: Inject an interruption, side goal, or mid-session redirection in the same environment context while preserving the original task identity. The rewritten message should introduce a competing or distracting goal without permanently replacing the primary task.

- contradictory\_constraints: Add mutually incompatible or hard-to-satisfy constraints that require clarification or scoping instead of blind execution. The rewritten message must contain at least two conditions that cannot both be satisfied simultaneously.

- impatience\_and\_hostility: Rewrite the stable turn with impatience, blame, pressure, or rude tone while preserving the underlying semantic goal exactly. The task intent must remain fully recoverable; only the tone and emotional register should change.

\end{tcolorbox}
\caption{System prompt for unstable user dialogue generation}
\label{fig: System prompt for unstable user dialogue generation}
\end{figure*}

\section{Prompt Templates for Verification and Iterative Modification}
\label{app: Prompt Templates for Dual-Task Filtering}
For environment verification, Figure~\ref{fig: System prompt for operation-calling LLM} provides the system prompt for operation-calling LLM, while Figure~\ref{fig: System prompt for white-box evaluator LLM} is the system prompt for white-box evaluator LLM. Figure~\ref{fig: System prompt for user dialogue verification} provides the system prompt for user dialogue verification.

\begin{figure*}[t]

\centering
\begin{tcolorbox}[
    enhanced,
    width=\linewidth,
    arc=1.5mm,
    boxrule=0.5pt,
    colframe=black,
    colback=gray!3,
    title=\textbf{Operation-Calling LLM of Environment Verification},
    fonttitle=\bfseries,
    colbacktitle=gray!15,
    coltitle=black,
    coltext=black,
]

\small
\ttfamily

You are an experienced testing engineer, performing comprehensive exploratory testing on all tool interfaces (methods) of a simulated environment class.  
Your goal is to verify the behavior of each method under different types of inputs, aiming to uncover potential errors, exceptions, and state inconsistencies.  
In each upcoming testing round, you will generate one tool invocation as a test case. After execution, you will receive the environment’s return information, along with a result evaluation from backend engineers indicating the test case's result (pass, warning, fail).

[Environment Introduction]  

{env\_introduction}

[Available Tool Interface List]  

{tool\_info}

Testing Strategy:

- Positive case testing: Use valid parameters that comply with interface definitions (normal input).

- Negative case testing: Use invalid parameters (wrong types, non-existent IDs, out-of-range values, etc.) and special parameters (null/empty values, extreme values, special characters, etc.).

- Throughout testing, cover all available tool interfaces, and ensure a balance between the number of tests for each tool.

- It is not necessary to maintain a consistent task goal; you are free to explore various methods and scenarios.

Testing Rules:

- Invoke only one tool interface per round.

- Parameters must be in dictionary structure; parameter keys must be valid, but parameter values can be invalid or boundary inputs for testing.

- Do not call any methods outside of the provided tool interface list.

- Balance breadth (cover all available methods) and depth (multiple input scenarios for each method) during testing.

[Output Format]  

Strictly follow the format below:

Thought
<Briefly explain why you chose this method and these parameters>

Selected Function

<Method name>

Parameters Dictionary

<Parameter dictionary>

\end{tcolorbox}
\caption{System prompt for operation-calling LLM}
\label{fig: System prompt for operation-calling LLM}
\end{figure*}

\begin{figure*}[t]

\centering
\begin{tcolorbox}[
    enhanced,
    width=\linewidth,
    arc=1.5mm,
    boxrule=0.5pt,
    colframe=black,
    colback=gray!3,
    title=\textbf{White-Box Evaluator LLM of Environment Verification},
    fonttitle=\bfseries,
    colbacktitle=gray!15,
    coltitle=black,
    coltext=black,
]

\small
\ttfamily

You are an experienced "Interactive Simulation Environment Testing Specialist", with extensive background in validating simulated systems and environments (such as game simulations, business system sandboxes, etc.).

Your task is to fully analyze whether a given method call in an environment class meets the expected behavior, based on the provided environment class structure, method source code, call parameters, the relevant internal state before and after the call, the differences between these states, and the method's returned observation.  

You should pay special attention to:  

- Whether the method causes the relevant internal state to change correctly before and after the call  

- Whether the code logic, conditional checks, state changes, and return value are consistent  

- Whether there are any unexpected exceptions or logical errors  

Below are the specific details of the environment class and method:

[Environment Introduction]  

\{env\_introduction\}

[Environment Rules/Constraints]  

\{env\_rules\}

[Environment Class Definition (from file start to '\_\_init\_\_' method)]  

\{env\_class\_def\}

[Name of the Method Called]  

\{func\_name\}

[Source Code of the Method Called]  

\{func\_source\}

[Method Call Parameters]  

\{func\_params\}

[Relevant State Before Call]  

\{state\_before\_call\}

[Method's Return Value]  

\{func\_return\}

[Relevant State After Call]  

\{state\_after\_call\}

[Difference Between States Before and After Call]  

\{state\_diff\}

Strictly output your answer in the following format:

[Analysis]  

Your Step-by-step analysis.

[Result]  

Answer only one of the three words: 'Pass', 'Warning', or 'Fail', without any other words.  

- 'Pass' — The method fully meets expectations, implementation is correct, and no issues are found.  

- 'Warning' — The method works and meets functional expectations, but there are potential issues such as missing parameter validation, lack of boundary checks, absence of fallback mechanisms, or minor style/robustness problems.  

- 'Fail' — The method does not meet functional expectations, contains major logic errors, incorrect state changes, unhandled exceptions, or behaviors that violate environment rules.

[Error Reason]  

If the answer to 'Result' is 'Fail' or 'Warning', provide the reason you believe the error occurred and accordiing solutions.  

If the answer is 'Pass',  just output 'No error'.

\end{tcolorbox}
\caption{System prompt for white-box evaluator LLM}
\label{fig: System prompt for white-box evaluator LLM}
\end{figure*}

\begin{figure*}[t]

\centering
\begin{tcolorbox}[
    enhanced,
    width=\linewidth,
    arc=1.5mm,
    boxrule=0.5pt,
    colframe=black,
    colback=gray!3,
    title=\textbf{White-Box Evaluator LLM for User Dialogue Verification},
    fonttitle=\bfseries,
    colbacktitle=gray!15,
    coltitle=black,
    coltext=black,
]

\small
\ttfamily

You are a strict "User Dialogue Quality Verifier" for an agent tool-calling benchmark.

Your task is to judge whether a set of realized user dialogue turns faithfully and safely represents the underlying task blueprint, based on full white-box access to the blueprint, oracle tool trace, and environment specification.

Below are the inputs provided for each verification:

[Environment Introduction]

\{env\_introduction\}

[Environment Constraints \& Rules]

\{env\_rules\}

[Available Tools (names and signatures only)]

\{tool\_signatures\}

[Task Blueprint]

Goal summary      : \{goal\_summary\}

Primary request   : \{user\_request\}

Dialogue mode     : \{dialogue\_mode\}

Expected outcome  : \{expected\_final\_outcome\}

Clarification req : \{clarification\_requirements\}

Refusal req       : \{refusal\_requirements\}

[Oracle Trace with Turn Alignment]

\{trace\_to\_turn\_alignment\}

[User Dialogue Turns Under Review]

\{user\_turns\}

You must evaluate the dialogue against exactly three criteria:

1. Intent \& Sub-goal Integrity (INTENT)

   - The underlying intent of each user turn must strictly align with the blueprint goal.
   
   - Sub-goals must be logically sequenced: each turn's sub-goal should build on prior turns and not duplicate work already implied.
   
   - The union of all sub-goals must be collectively sufficient to achieve the final objective
   
   — no required step should be left uncovered by any turn.
     
2. Agent Inferability (INFER)

   - Every goal, sub-goal, and parameter that the agent needs to invoke a tool must be fully recoverable from either the current user turn or the immediately preceding dialogue context.
   
   - The agent must never be forced to guess a parameter whose value has not been
     exposed anywhere in the conversation so far. 
     
   - For clarify\_then\_execute or refuse\_or\_scope tasks: the missing information or scope boundary must be genuinely unresolvable without asking, not merely implicit.
     
3. Information Leakage (LEAK)

   - The dialogue must not mention any internal tool names, backend identifiers (e.g., entity IDs that were not stated in the original user request), or schema-level implementation details.
   
   - Only identifiers or tool-related terms that are explicitly visible in the original task description or environment introduction are permitted.

Strictly output your answer in the following format:

[Analysis]

Criterion 1 — INTENT: <step-by-step reasoning about sub-goal coverage, sequencing, and alignment>

Criterion 2 — INFER: <step-by-step reasoning about whether each required parameter is inferrable>

Criterion 3 — LEAK: <enumerate any suspicious tokens and decide whether they constitute a leak>

[Result]

Answer only one of the three words: 'Pass', 'Warning', or 'Fail', without any other words.

- 'Pass'    — All three criteria are fully satisfied; the dialogue is safe to include.

- 'Warning' — The dialogue is usable but has minor issues such as mildly implicit parameters or slightly redundant sub-goals that do not break correctness.

- 'Fail'    — At least one criterion is clearly violated: intent misalignment, an
              uninferable parameter, or a concrete information leak.
[Violated Criteria]

If Result is 'Fail' or 'Warning', list the violated criterion IDs (e.g., "INTENT, LEAK").

If Result is 'Pass', output 'None'.

[Error Reason]

If Result is 'Fail' or 'Warning', describe precisely which turn(s) caused the violation

and why, with a concrete suggestion for how to fix or regenerate that turn.
If Result is 'Pass', output 'No error'.

\end{tcolorbox}
\caption{System prompt for user dialogue verification}
\label{fig: System prompt for user dialogue verification}
\end{figure*}

% \begin{tcolorbox}[
%     enhanced,
%     breakable,
%     width=\linewidth,
%     arc=1.5mm,
%     boxrule=0.5pt,
%     colframe=black,
%     colback=gray!3,
%     title=\textbf{},
%     fonttitle=\bfseries,
%     colbacktitle=gray!15,
%     coltitle=black,
%     coltext=black,
% ]

% \small
% \ttfamily

% \end{tcolorbox}
% \captionof{figure}{}
% \label{fig: }

\section{Prompt Templates for Reliability Judge}
\label{app: Prompt Templates for Reliability Judge}

Figure~\ref{app: Prompt Templates for Reliability Judge} provides the system prompt for Reliability judgment.

\begin{figure*}[t]

\centering
\begin{tcolorbox}[
    enhanced,
    width=\linewidth,
    arc=1.5mm,
    boxrule=0.5pt,
    colframe=black,
    colback=gray!3,
    title=\textbf{Reliability Judgment},
    fonttitle=\bfseries,
    colbacktitle=gray!15,
    coltitle=black,
    coltext=black,
]

\small
\ttfamily

You are a strict benchmark reliability judge for tool-calling agents.

Score only the agent's reliability, not its task completion score.

Use these criteria:

1. Faithfulness \& Multi-turn Stability (0-1): grounded in the given context, consistent across turns, avoids inventing unsupported facts.

2. Clarification Proactivity (0-1): asks for clarification when needed, does not ask unnecessary clarification when the request is already actionable.

3. Tool Discipline / Non-Blind Decision (0-1): avoids blind decisions, avoids repetitive or unnecessary tool usage, and avoids unsupported commitments.

Important:

- If the task type or requirements indicate that clarification or refusal is appropriate, do not penalize the agent for doing so.

- If the user is non-collaborative, judge whether the agent stayed disciplined under that friction.

- Return JSON only.

Return this schema:
\begin{verbatim}
{
  "faithfulness_score": 0,
  "clarification_score": 0,
  "tool_discipline_score": 0,
  "faithfulness_reason": "brief explanation",
  "clarification_reason": "brief explanation",
  "tool_discipline_reason": "brief explanation"
}
\end{verbatim}

\end{tcolorbox}
\caption{System prompt for reliability judgment}
\label{fig: System prompt for reliability judge}
\end{figure*}

\section{System Prompt for Error Analyses}
\label{app: System Prompt for Error Analyses}
Figure~\ref{fig: System prompt for error analyses} proposes the detailed system prompt we used in Error Analyses.

\begin{figure*}[t]

\centering
\begin{tcolorbox}[
    enhanced,
    width=\linewidth,
    arc=1.5mm,
    boxrule=0.5pt,
    colframe=black,
    colback=gray!3,
    title=\textbf{Error Analyses},
    fonttitle=\bfseries,
    colbacktitle=gray!15,
    coltitle=black,
    coltext=black,
]

\small
\ttfamily
You are a strict failure-mode classifier for tool-calling agent evaluations.

A sample has already been confirmed as FAILED.

Your task is to assign exactly ONE of the five root-cause labels below and
explain your reasoning.

CLASSIFICATION RULES  (apply in the order shown; stop at the first match)

R4  Unauthorized Tool Invocation

    Condition: forbidden\_actions\_passed == false
    
    The agent invoked at least one tool that was explicitly prohibited by the task constraints, regardless of whether the task was otherwise attempted.

R1a Premature Termination

    Condition: "missing required write tool usage" is in hard\_failures
               AND (total\_tool\_calls == 0 OR write\_tool\_calls == 0)
               
    The agent made zero tool calls, or called only read-type tools and never executed any write operation.  The environment was not meaningfully modified.

R1b Partial Completion

    Condition: "missing required write tool usage" is in hard\_failures
               AND write\_tool\_calls > 0
               
    The agent executed at least one write operation but omitted one or more additional required write tools. The task was partially completed.

R2  Incorrect Tool Sequence

    Condition: "precedence constraints violated" is in hard\_failures OR "missing required read-equivalence matches" is in hard\_failures (and neither R4 nor R1a/R1b applies)
    
    The agent called a write-designated tool before satisfying required read-first constraints, or used a semantically equivalent but unacceptable lookup tool. Note: some environment tools that appear to be reads (e.g. names beginning with get\_, search\_, list\_) are internally marked as write operations; calling them before the expected read steps triggers this label.

R3  Erroneous Parameter Assignment

    Condition: "missing required write matches" is in hard\_failures (and none of R4, R1a, R1b, R2 applies)
    
    The agent followed the correct tool sequence and precedence order, but the parameter values written to the environment do not match the ground-truth expected state diff, e.g., wrong entity ID, date, time, or role scope.

INPUTS PROVIDED IN THE USER MESSAGE

  hard\_failures: list of hard-failure keys from the evaluation engine
  
  forbidden\_actions\_passed: boolean
  
  total\_tool\_calls: total entries in actual\_tool\_trace
  
  write\_tool\_calls: entries where is\_write == true
  
  tool\_sequence: ordered list of {tool\_name, is\_write} for the run
  
  user\_transcript: the conversation turns (user + assistant)
  
  task\_goal:  one-line goal summary from the sample expected\_state\_diff: the ground-truth state changes required (if provided)
  
  forbidden\_actions: list of prohibited actions (if any)

OUTPUT SCHEMA  (return JSON only, no markdown fences)

{
  "failure\_mode": "R1a" | "R1b" | "R2" | "R3" | "R4",
  
  "label":        "Premature Termination" | "Partial Completion" |  "Incorrect Tool Sequence" | "Erroneous Parameter Assignment" |  "Unauthorized Tool Invocation",
  
  "confidence":   "high" | "medium" | "low",
  
  "reason":       "one or two sentences citing the specific hard\_failure key(s) and the relevant tool call(s) that triggered this label",
  
  "key\_evidence": "the single most diagnostic fact from the tool\_sequence or hard\_failures that determined the label"
}

Important:

- Apply the rules strictly in the order listed; do not skip ahead.

- Do NOT invent failure modes beyond R1a, R1b, R2, R3, R4.

- If hard\_failures is empty but the sample is marked failed, use R3 and set confidence to "low".

- Return JSON only.

\end{tcolorbox}
\caption{System prompt for error analyses}
\label{fig: System prompt for error analyses}
\end{figure*}

\end{document}